\documentclass{article}

\PassOptionsToPackage{compress}{natbib}
\usepackage{amsmath}
\usepackage{subfig}
\usepackage{booktabs} %
\usepackage{multicol} %
\usepackage{array, makecell} %

\usepackage[preprint]{neurips_2021}

\usepackage[utf8]{inputenc} %
\usepackage[T1]{fontenc}    %
\usepackage{url}            %
\usepackage{booktabs}       %
\usepackage{amsfonts}       %
\usepackage{nicefrac}       %
\usepackage{microtype}      %
\usepackage{xcolor}         %
\definecolor{MK_One_One}{RGB}{43,140,190}
\definecolor{MK_Two_Two}{RGB}{178,24,43}

\usepackage[colorlinks]{hyperref}
\hypersetup{               
    linkcolor=MK_Two_Two,
    citecolor=MK_One_One,
    urlcolor=MK_Two_Two,
}

\usepackage{enumerate, amsmath, amsthm, amssymb, pifont, csquotes, dashrule, mathrsfs, tikz, bbm, booktabs, bm, verbatim, cases, algorithm,  algorithmic}
\usepackage[framemethod=TikZ]{mdframed}
\usepackage{url}
\usepackage{amsmath}
\usepackage{enumitem}

\usepackage{wrapfig}
\usepackage{setspace}
\usepackage{hyperref}

\newcommand{\AF}{{{\mathcal{A} \mathcal{F}}}}

\newcounter{KDefCounter}

\newcommand{\ddef}[2]
{
\vspace{1mm}
\refstepcounter{KDefCounter}
{\bf Definition \theKDefCounter} (#1): {\it #2}
}
\usepackage{hyperref}

\makeatletter
\newcommand{\printfnsymbol}[1]{%
  \textsuperscript{\@fnsymbol{#1}}%
}
\makeatother

\newcommand*{\affaddr}[1]{#1} %
\newcommand*{\affmark}[1][*]{\textsuperscript{#1}}

\usepackage{blindtext} 


\title{The Paradox of Choice: Using Attention in Hierarchical Reinforcement Learning}

\author{%
 Andrei Nica\thanks{equal contribution}~ \affmark[1,2,3] \and \textbf{Khimya Khetarpal\printfnsymbol{1}\affmark[1,2]} \and \textbf{Doina Precup}\affmark[1,2,4] \\
\affaddr{\affmark[1]McGill University}, \affaddr{\affmark[2]Mila}, \affaddr{\affmark[3]Polytechnic University of Bucharest}, \affaddr{\affmark[4]Google Deepmind} \\
\texttt{khimya.khetarpal@mail.mcgill.ca, nicaandr@mila.quebec, dprecup@cs.mcgill.ca} \\
}

\usepackage{selectp}

\begin{document}

\maketitle

\begin{abstract}
Decision-making AI agents are often faced with two important challenges: the depth of the planning horizon, and the branching factor due to having many choices. Hierarchical reinforcement learning methods aim to solve the first problem, by providing shortcuts that skip over multiple time steps. To cope with the breadth, it is desirable to restrict the agent's attention at each step to a reasonable number of possible choices. The concept of affordances (Gibson, 1977) suggests that only certain actions are feasible in certain states. In this work, we model ``affordances'' through an attention mechanism that limits the available choices of temporally extended options. We present an online, model-free algorithm to learn affordances that can be used to further learn subgoal options. We investigate the role of \textit{hard} versus \textit{soft} attention in training data collection, abstract value learning in long-horizon tasks, and handling a growing number of choices. We identify and empirically illustrate the settings in which the paradox of choice arises, i.e. when having fewer but more meaningful choices improves the learning speed and performance of a reinforcement learning agent.	\end{abstract}

\section{Introduction}
\label{sec:intro}
Decision making in complex environments can be challenging due to having many choices to consider at every time step. Learning an attention mechanism that limits these choices could potentially result in much better performance. Humans have a remarkable ability to selectively pay attention to certain parts of the visual input~\citep{judd2009learning, borji2012salient}, gathering relevant information, and sequentially combining their observations to build representations across different timescales~\citep{hayhoe2005eye, zhang2019atari}, which plays an important role in guiding further perception and action~\citep{nobre2019premembering, badman2020multiscale}. In this paper, we explore ideas for endowing reinforcement learning (RL) agents with these type of capabilities.

A RL agent interacts with an environment through a sequence of actions, learning to maximize its long-term expected return~\citep{sutton2018introduction}. Temporal abstraction, e.g. options~\citep{sutton1999between} allow it to consider decisions at variable time scales. %
Options allow the agent to reduce the depth of its lookahead when performing planning, and to propagate credit over a longer period of time. However, in large state-action spaces, the agent is still faced with many choices at every decision step, and options can in fact worsen this problem, as choices multiply when considering different timescales. Hence, temporal abstraction can lead to a larger branching factor.

A well-known solution is to {\em select} a small number of choices to focus on, for instance, applying an action only when certain preconditions are met in classical planning~\citep{Fikes1972}, or using initiation sets for options~\citep{sutton1999between}. Recent work~\citep{khetarpal2020options} proposed a generalization of initiation sets to \textit{interest functions}~\citep{Sutton2016,White17}, which provide a way to learn options that specialize to certain regions of the state space. All these approaches can be viewed as providing an attention mechanism over action choices.

While attention has been widely studied in the field of computer vision~\citep{hou2007saliency, borji2012salient}, the role of different types of attention over action choices has not been studied much, especially for temporally abstract actions. In this work, we explore attention over action choices in RL agents with a special interest in affordances associated with options, which can be viewed as a form of \textit{hard} attention. On the other hand, having soft preferences over \textit{all} actions without eliminating any can be viewed as \textit{soft} attention. 

In this paper, we study the role of \textit{hard} versus \textit{soft} attention in decision making with temporal abstraction. We posit that in certain settings, restricting an agent's attention through affordances leads to fewer but more useful choices compared to using soft attention in the form of interest functions. We demonstrate empirically this \textit{the paradox of choice:} fewer choices can contribute to faster learning, resulting in more rewards. As expected, this effect is more pronounced when the agent attends to choices that lead to better long-term utility. Attending to useful choices in a given state often has a compound effect, leading to further good choices in the near future.

\textbf{Key Contributions:} We measure the impact of \textit{hard} versus \textit{soft} attention over option choices: 1) when generating training data, 2) on abstract value learning in long-horizon tasks, and 3) when increasing the number of choices. We present an online, model-free algorithm to learn affordance-aware subgoal options and the policy over options with given subgoals and a high-level task. The affordances and option policies are learned simultaneously, informing each other. %
Leveraging the learned affordances to generate training data facilitates sample-efficient learning of the options. We empirically demonstrate that fewer choices can be better for both learning speed and final performance for an hierarchical reinforcement learning agent.

The focus of our work is to study affordances as hard attention and contrast it with soft-attention for temporally extended actions. We demonstrate this through non-tabular environments, consisting of both discrete and continuous action/state space in Minigrid~\citep{gym_minigrid} %
and Open-AI Robotics~\citep{schulman2017proximal} respectively.
The choice of these experimental setups were motivated by careful design with the aim to investigate and isolate the role of affordances as hard attention, while being able to vary the complexity in terms of number of options and tasks. We also emphasize that our approach scales to both discrete and continuous action spaces and it can be used in any domain where subgoals are known apriori.

\section{Preliminaries}
\textbf{Markov Decision Process (MDP).} A finite, discrete time MDP is a tuple $M=\langle {\cal S}, {\cal A}, r, P \rangle $, where ${\cal S}$ is a finite set of states, ${\cal A}$ is a  finite set of actions, $r: {\cal S} \times {\cal A}\rightarrow \mathbb{R}$ is the reward function, $P:{\cal S} \times {\cal A} \rightarrow Dist({\cal S})$ is the  transition dynamics, mapping state-action pairs to a distribution over next states. At each time step $t$, the agent observes a state ${ S}_t \in {\cal S}$ and takes an action ${A}_t \in {\cal A}$ drawn from its policy $\pi$. A stochastic policy is a conditional distribution over actions given a state $\pi: {\cal S} \times {\cal A} \rightarrow [0,1]$ and a deterministic policy is a mapping $\pi: {\cal S} \rightarrow {\cal A}$. The infinite-horizon cumulative \textit{discounted} return is defined as: $G_t = \sum_{t=1}^{\infty} \gamma^{t-1} r(S_t,A_t)$, where $\gamma\in (0,1)$ is the discount factor. The value function for $\pi$ is defined as: $V_\pi(s) = \mathrm{E}_{\pi}[G_t | S_t =s]$ and action value function is: $Q_\pi(s,a) = \mathrm{E}_{\pi}[G_t | S_t =s, A_t = a]$. 

\textbf{Hierarchical Reinforcement Learning (HRL)} aims to find closed-loop policies that an agent can choose to use for some extended period of time, also known as \textit{temporally extended actions}.  Various HRL approaches have been proposed~\citep{dayan1993feudal, thrun1995finding, parr1998reinforcement, dietterich2000hierarchical, vezhnevets2017feudal, nachum2018data}. Our work uses the options framework~\citep{sutton1999between}, because it is general, and it also connects well with the notion of attention over actions, through the concept of initiation sets. 

A Markovian option~\citep{sutton1999between} $\omega \in \Omega$ is composed of an \textit{intra-option policy} $\pi_\omega$, a termination condition $\beta_\omega: {\cal S} \rightarrow [0,1]$, where $\beta_\omega(s)$ is the probability of terminating the option upon entering state $s$, and an initiation set $I_\omega \subseteq {\cal S}$. In the \textit{call-and-return} option execution model, when an agent is in state $s$, it first examines the options that are available, i.e., for which $s\in I_\omega$. Let $\Omega(s)$ denote this set of available options. The agent then chooses $\omega\in \Omega(s)$ according to the policy over options $\pi_\Omega(s)$, follows the internal policy of $\omega$, $\pi_\omega$, until it terminates according to $\beta_\omega$, at which point this process is repeated. This execution model defines a \textbf{semi-Markov decision process (SMDP)}~\citep{puterman1994markov}, in which the amount of time between two decision points is a random variable depending on the state-option pairs. Options can be learned by defining  subgoal states and positively rewarding the agent for reaching them~\citep{sutton1999between, mcgovern2001automatic, vezhnevets2017feudal}, in which case they are called~\textbf{subgoal options}. 

\textbf{Affordances.} Affordances are a property of the agent and its environment~\citep{gibson1977theory, heft1989affordances, chemero2003outline}, indicating that only certain actions are feasible given certain features of the state. Affordances have been typically formalized in context of the objects~\citep{slocum2000further, fitzpatrick2003learning, lopes2007affordance, montesano2008learning, cruz2016training, cruz2018multi}, e.g. a chair affords sitting, a cup affords grasping, etc. In classical AI systems, affordances have been  viewed as preconditions~\citep{Fikes1972} for actions. Our view of affordances is most aligned with goal-based priors~\citep{abel2014toward, abel2015goal}. In particular, we build upon the notion of action affordances defined via intents~\citep{khetarpal2020i}. Intents are defined as a target state distribution that should be achieved after the execution of an action, thereby defining a notion of action success. We will extend this concept to options, by considering intent achievement as a property that also depends on time, similarly to work on empowerment~\citep{salge2014empowerment} or on goal conditioning~\citep{schaul2015uvfa,nachum2018data}.

\section[Affordances as Hard Attention]{Choice Attention for Reinforcement Learning}
\label{sec:hardattention}

\begin{wrapfigure}{R}{0.5\textwidth}
    \begin{minipage}{0.5\textwidth}
        \begin{figure}[H]
          \vspace{-0.8cm}
          \centering
            \includegraphics[width=1.0\textwidth]{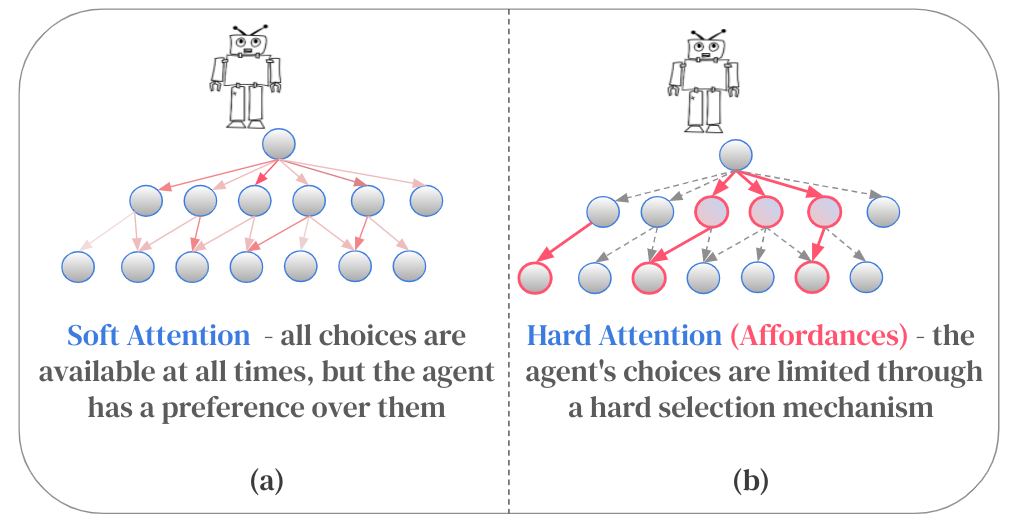}
            \caption{%
           Soft attention vs hard attention over action (or option) choices. Soft preferences are indicated by different weights of arrows (a). Affordances, like hard attention, allow only a subset of choices, represented as solid arrows (b)}\vspace{-0.3cm} \label{fig:attentiontypes}
        \end{figure}
    \end{minipage}
\end{wrapfigure}

Consider the agent in Fig.~\ref{fig:attentiontypes}, which must choose from many different action possibilities at every state. In value-based methods, every choice involves computing the action-value function of all options, in order to infer which one is best. This process can be expensive, if there are many action choices. Moreover, the values of different actions are imperfect estimates, and if incorrect, they can mislead the agent into a poor decision. In temporal abstraction, these estimation errors can affect options unevenly, since they can depend on other factors that affect variance, such as the duration of the options or the termination condition. Hence, comparing options of different duration can be problematic. Using policy-based methods may appear to be a solution, but these methods still leverage value estimates in training, so the problem described still exists (albeit hidden). Finally, agents that do explicit planning using a model may further suffer from having a large number of choices, because it explodes their search space.

Attention mechanisms in animals~\citep{hayhoe2005eye, tatler2011eye} as well as in computer vision~\citep{xu2015show, luong2015effective} and in language~\citep{brown2020language} can be used to avoid the effect of noise and improve performance. In HRL agents, attention over actions can provide a similar effect. We consider and compare two approaches for implementing attention in this context.

Soft-attention (Fig.~\ref{fig:attentiontypes}(a)) places ``soft'' preferences over action choices, but does not rule out any of them. Interest functions~\citep{khetarpal2020options} for options can be used to implement this idea. Appropriately tuned interest can reduce the noise in an agent's behavior. In contrast, an agent that understands which actions are affordable can use the information to strictly limit its attention to much fewer choices (Fig.~\ref{fig:attentiontypes}(b)). Consequently, affordances can be viewed as \textit{hard} attention~\citep{luong2015effective, xu2015show}. %
In model-free RL, this leads to a much narrower distribution of states from which to bootstrap, potentially lowering the variance of the value updates. The benefits further increase for planning, where option models allow the agent to reason ``deeper", by considering a longer horizon, while affordances can reduce breadth of the search tree.

The potential gains obtained by using affordances (i.e. hard attention) can further be understood through the \textbf{\textit{paradox of choice}}, an idea due to Herbert Simon and popularized by \citet{schwartz2004paradox}.
The main idea is that having too many choices can be worse for decision making, because the optimization problem becomes harder. A related concept is that of finding ``satisficing" solutions for a problem, rather than fully optimizing it. A reduced number of choices can facilitate this process. While these ideas have been mainly discussed in the context of human decision making, our work can be viewed as providing a possible implementation and evaluation for artificial RL agents. Intuitively, HRL is itself an attempt to replace exact return maximization, with a solution that is sub-optimal but easier to compute, i.e., satisficing.

\section{Affordance-Aware Subgoal Options}
\label{sec:formalization}

In this section, we introduce \textit{affordance-aware subgoal options} in Sec.~\ref{sec:problemformulation}, and propose a method for learning them for an HRL agent in Sec.~\ref{sec:method}.

\subsection{Problem Formulation}
\label{sec:problemformulation}

Let $g\in{\cal S}$ be an arbitrary subgoal state, which is the desired target for an option. We assume for simplicity that subgoals are given\footnote{Note that many approaches have been proposed to learn sub-goals, and they could be used in an outer loop for our approach.}. Similarly to~\citet{khetarpal2020i}, we will have to define intents and affordances for options, with respect to sub-goals.

\ddef{Subgoal-option Intent Satisfaction}{A state $s$ satisfies intent $I_{\omega,g}$ to degree $\epsilon\in(0,1)$ if the probability of reaching $g$ while executing $\omega$ from $s$, before $\omega$ terminates, is larger than $\epsilon$. 
\label{def:subgoalintent}}

Consider trajectories $\tau$ generated by running $\omega$ from $s$ until completion. On any trajectory, consider an indicator function that marks if $g$ was achieved or not. The intent $I_{\omega,g}$ is then achieved if the expected value of this indicator function, which we call {\em intent completion function}, is above $\epsilon$.

\ddef{Option Affordances $\AF_{\cal G}$}{
Given option set $\Omega$, subgoal set ${\cal G} \subseteq{\cal S}$ and $\epsilon>0$, the affordance associated with ${\cal G}$, $\AF_{\cal G}\subseteq {\cal S} \times \Omega $, is a relation such that $\forall (s,\omega)\in \AF_{\cal G}$, state $s$ satisfies intent $I_{\omega,g}$ to degree $\epsilon\in(0,1)$
\label{def:optionaffordances}}

\ddef{Affordance-aware Subgoal Options}{Given a subgoal $g$, an affordance-aware subgoal option ${\omega \in \Omega}$ is composed of a tuple $\langle \AF_g(\omega), \pi_{\omega}(a|s), \beta_{\omega}(s) \rangle$, where $\AF_g(\omega)$ denotes the set of states that afford $\omega$ with respect to $g$, $\pi_{\omega}$ is the intra-option policy, and $\beta_{\omega}(s)$ is the termination condition.
\label{def:subgoaloptions}}

\begin{wrapfigure}[12]{r}[-0mm]{7cm}
    \begin{minipage}{0.5\textwidth}
        \begin{figure}[H]
            \vspace{-1.2cm} 
            \centering
            \subfloat[Domain\label{fig:pickupblueobj-domain}]{\includegraphics[width=0.28\textwidth]{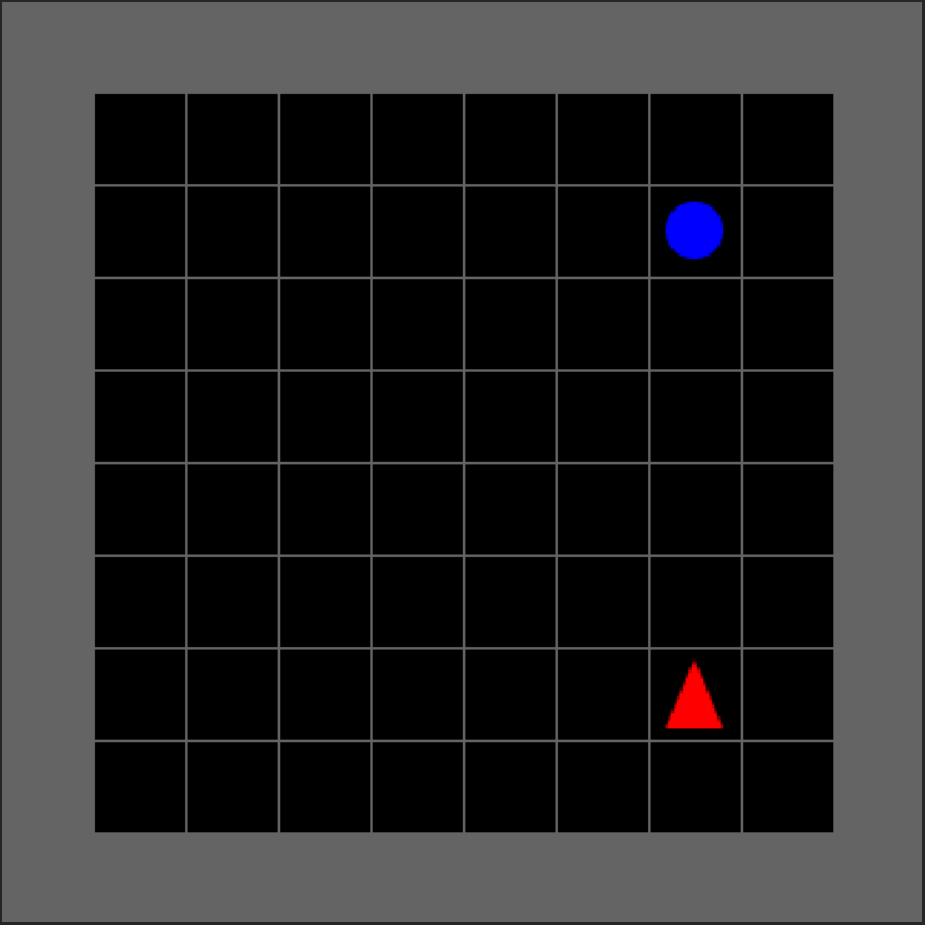}} \hspace{4mm} %
            \subfloat[Affordances \label{fig:pickupblueobj-affordances}]{\includegraphics[width=0.28\textwidth]{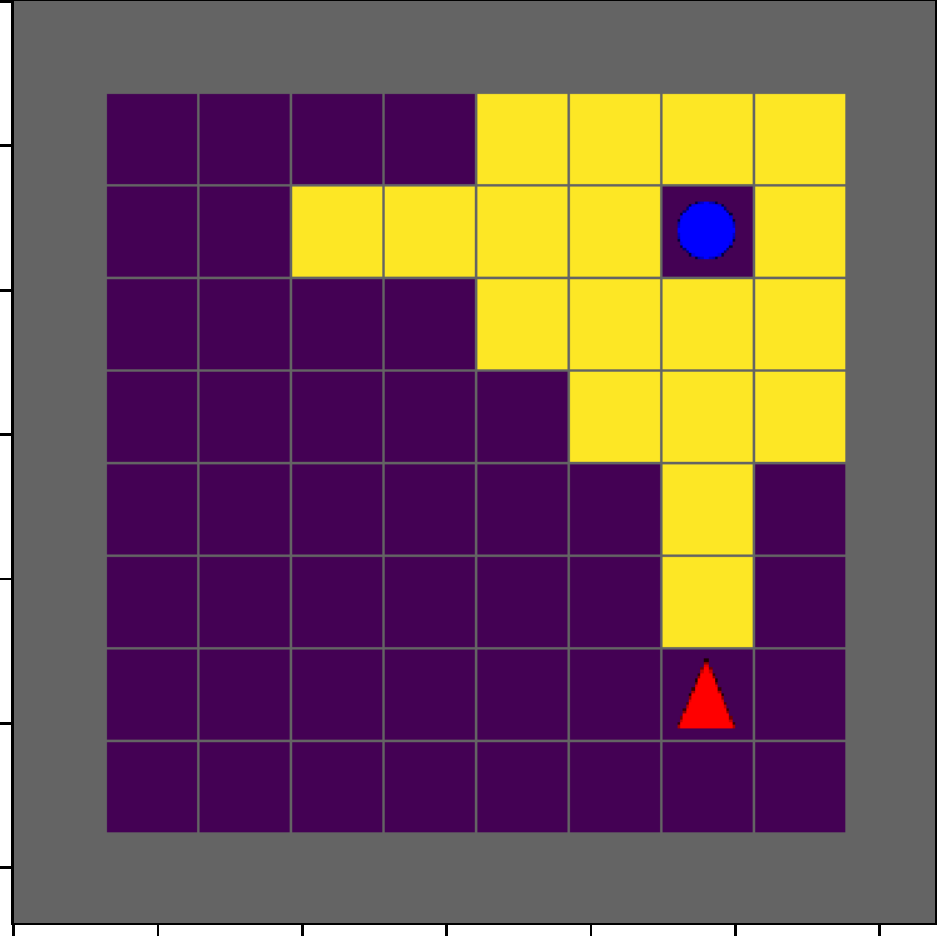}} %
            \vskip -0.15in
            \caption{\label{fig:illustrationaffordances} The environment and ground truth affordance for the subgoal-option ``pick-up-blue-object''. The option is affordable closer to the blue object and unaffordable in states further away.}
            
        \end{figure}
    \end{minipage}
\end{wrapfigure}

To illustrate these definitions, we consider the environment depicted in Fig.~\ref{fig:pickupblueobj-domain}, where an agent needs to navigate to specific objects, which it can then use to achieve certain tasks. Consider the intent of picking up the blue object in a maximum of 5 timesteps, and suppose that we have trained an option that can navigate to the blue object. Fig.~\ref{fig:pickupblueobj-affordances} shows the states that are in the affordance for this option with respect to the stated goal, indicated by yellow. The details of the affordance computation are discussed in the next section.

\subsection{Learning Algorithm}
\label{sec:method}
Suppose that a set of subgoals is provided to the agent. These subgoals can be associated with pseudo-rewards, and we can train options that attempt to maximize these. Simultaneously,the corresponding affordances can be constructed by identifying the states in which the options achieve their intended goal. We now present the details of this process.

\textbf{Core Idea.}
For a given option $\omega \in \Omega$ and subgoal $g$, the intent completion function takes a trajectory generated by the option's internal policy, $\pi_\omega$, and returns a binary value, indicating whether the desired subgoal is reached on that trajectory. %
This offers a target to learn the affordance $\AF_{\cal G}(\omega)$. To facilitate learning,  we would like the intent completion to reflect \textit{gradual} change, as the agent approaches the completion of the desired subgoal. %
To model this temporally extended aspect, we use a \textit{discounted}-intent completion function ($\texttt{IC}$) with discount $\gamma_{\texttt{IC}} \in [0, 1)$. Note that this discount factor is different than the one used to learn option policies. %
Moreover, the intent completion function measures the completion for all partial trajectories of an option: for trajectory $\tau =\{ s_{0}, s_{1}, \dots, s_{t} \}$, all partial trajectories $\{ s_{0}, s_{1} \}, \{ s_{0}, s_{1}, s_{2} \}, \{ s_{0}, \dots, s_{t} \}$ are evaluated for intent completion, resulting in the following target: $\texttt{IC}^{target}(s_{t},\omega_t)=$
\begin{equation}
    \begin{aligned}
        \begin{cases}
        \gamma_{\texttt{IC}} \times \texttt{IC}^{target}(s_{t+1},\omega_t), & \text{if }\mathbbm{1}_g(\tau_{s_{1}:s_{t}})=0 
        \\
        \mathbbm{1}_g(\tau_{s_{1}:s_{t}}),  & \text{otherwise}
        \end{cases}
    \end{aligned}
\label{eq:affordancetarget}
\end{equation}

\textbf{Data Generation.} The key idea to learning subgoal options in an online, model-free fashion is for the agent to generate a training dataset of option trajectories labelled with intent completion targets (Eq.~\ref{eq:affordancetarget}). This is challenging due to the \textit{chicken-and-egg} problem of knowing which option to sample, \textit{while} the options are being learned. In applications where a wealth of data is available, our approach could leverage expert trajectories, as in imitation learning, or human-in-the-loop learning. However, here we deal with a more challenging, online scenario. %
To cope with online data generation, %
we investigate the role of different attention mechanisms, namely, uniformly random, soft, and hard attention. It is important to note that the attention is learned simultaneously (through Eq.~\ref{eq:affordancetarget}). This poses an additional challenge: in the initial stages of learning, the attention values are almost random. To better understand this, we further investigate if knowing which options to sample facilitates learning of options themselves (see  Sec.~\ref{sec:sampling}). 

\begin{algorithm}[]
   \caption{\texttt{DataGeneration}}
   \label{alg:dataset_trajectoryoptions}
\begin{algorithmic}
   \STATE {\bfseries Require:} Intents $\mathcal{I}$, Environment $E$, Maximum trajectory length, $n_{max}$, Number of parallel environments $nenvs$, Steps per environment $nsteps$.
   \STATE {\bfseries Require:} Intent completion $I_{\omega, g}$, $\texttt{IC}^{net}_{z}, \pi_{\omega,\theta}, \beta_{\omega,\theta}$
   \STATE {\bfseries Require:} $\mathcal{D}_{\tau}$ - Dataset of completed trajectories, $\mathcal{B}_{\tau}$ - Buffer of trajectories
   \FOR{$i=1 \dots  nenvs$}
   	 \STATE Load $\tau$ from buffer $\mathcal{B}_{\tau}$ for env $i$
      \FOR{$t=1 \dots nsteps$}
        \STATE Sample $s \sim E$ 
        \STATE $\omega \leftarrow \omega_{\tau_{i}[t-1]}$ if $t > 1$ else $f_{att}(s)$
      	 \STATE Take $a$ according to $\pi_{\omega,\theta}(a|s)$ and observe $s', r$
     	 \STATE $\{(s, \omega, a, s', \beta_\omega(s')), \mathbbm{1}_{g_{w}}(\tau_{i, s_{1}:s_{t}}) \} \rightarrow$ append to $\tau_{i}$
    	 \IF{any intent is completed \OR $|\tau_{i}| \geq n_{max}$}
    		\STATE Sample $\omega$ at $s$ according to $f_{att}$ using Alg.~\ref{alg:attention-modelfree}.
            \STATE $\mathcal{D}_{\tau}\leftarrow  \{(\tau_{i})\}$ Add completed trajectory
            \STATE Reset trajectory buffer  $\tau_{i}$
        \ENDIF
      \ENDFOR
   \STATE $\mathcal{D}\leftarrow \cup_{\omega \in \Omega} \{(\tau)\}_{n_{max}}$. Collect all transitions
   \ENDFOR
\end{algorithmic}
\end{algorithm}

\textbf{Attention Mechanism.} We train a convolution neural network (see Sec.~\ref{experiments} for details), referred to as $\texttt{IC}^{net}$, which takes as input a state and option id and predicts the discounted intent completion ($\texttt{IC}^{target}$). To obtain \textit{hard}-attention, i.e. \textbf{\textit{affordances}}, we consider a \textit{hard}-threshold on the predicted value. The threshold value is a hyper-parameter which controls the number of affordable options, based on the distance to the corresponding subgoal, i.e. subgoal reachability. A threshold of $k$ means that values of discounted intent completion lower than $\gamma_{IC}^k$ are replaced by $0$s, sparsifying the set of options that the agent considers. The effect is two-fold: 1) options which are further from completing their associated subgoal are deemed less affordable and 2) the option duration is in effect controlled according to the choice of subgoal-intent specification. With a well designed choice of intents, obtaining affordances by thresholding can significantly reduce the size of the policy class used by the agent, thereby reducing the problem complexity, especially in large action spaces. This approach relies on having good subgoals, as they ultimately induce both the option policies and the intents. In this paper, we rely on domain knowledge to generate good subgoals. %

\textbf{Choice of Threshold \& Sensitivity.} We chose the threshold $k$ by analyzing the normalized predicted values of $\texttt{IC}^{net}$. We observed that a hard-threshold of $k=90$ separates the two distributions of affordable and non-affordable subgoals with high accuracy. Thus, we use this value of threshold for all our experiments. We provide more details in the App~\ref{sec:appendix-attentionAnalysis}. Next, We investigate the sensitivity to varying values of $k\in\{(5, 15, 90, 140\}$. The results corroborate the intuition that lower values of $k$ (e.g. 5, 15) lead to fewer affordable options, including no affordable options for states farther in time (due to our design of affordances via discounted intent). However, larger values of $k$ (e.g 90, 140) result in more options being affordable, which facilitates generating option trajectories with better coverage of the state space. We demonstrate a detailed evaluation of threshold values, $\texttt{IC}^{net}$ predictions and training sensitivity to them in the App~\ref{sec:appendix-attentionAnalysis}.

To obtain \textit{soft}-attention, i.e. \textbf{\textit{interest}}, we compute a softmax over $\texttt{IC}^{net}$ predictions for all options given a state. While our implementation of interest over option choices (i.e. soft-attention) is closely related to the interest function introduced in interest-option-critic (\citealp[IOC,][]{khetarpal2020options}), there are two key differences: 1) soft attention in our approach is learned using downstream-task-agnostic intent completion targets, as opposed to a task-specific reward function in IOC, and 2) we use a softmax over predicted discounted intent completion values, instead of gradient-based updates to parameterized interest as in IOC. Alg.~\ref{alg:attention-modelfree} shows the computation of the attention mechanisms. We note here that we used the default softmax function to obtain the soft attention (see Alg~\ref{alg:affordanceawareoptions-modelfree}). One could also consider softmax with temperature function, as it facilitates transitioning from soft-attention to hard attention. See detailed analysis in the App~\ref{sec:appendix-softtohard}.

\begin{algorithm}[]
		\caption{Attention Mechanism $f_{att}$}
		\label{alg:attention-modelfree}
		\begin{algorithmic}
		    \STATE {\bfseries Require:} $\texttt{IC}^{net}_{z}$, Attention Type, Threshold $k$, $\gamma_{\texttt{IC}}$.
            \STATE $X=(X_{0}, ..., X_{n})$ and $X_{i}= \texttt{IC}^{net}_{z}(s, \omega_i)$
		    \IF{attention == \texttt{\textcolor{blue}{hard}}}
    		  \STATE \textbf{if} $X_i \geq \gamma^k_{\texttt{IC}}$ \textbf{or} ${max}(X) < \gamma^k_{\texttt{IC}}$ 
                    \textbf{then} {$\AF_{\cal G}(X_i) \leftarrow 1$}
                    
                    \textbf{else} {$\AF_{\cal G}(X_i) \leftarrow 0$} ; \textbf{Return} $\AF_{\cal G}$ 
            \ELSIF{attention == \texttt{\textcolor{blue}{soft}}}
                \STATE $\texttt{Interest} \leftarrow \texttt{Softmax}(X)$ ; \textbf{Return} $\texttt{Interest}$  
            \ELSE
                \STATE $\texttt{Uniform}(X_i) \leftarrow 1$ ; \textbf{Return} $\texttt{Uniform}$  
    	     \ENDIF
		\end{algorithmic}
\end{algorithm}

\begin{algorithm}[]
		\caption{Model-Free Subgoal Option Learning}
		\label{alg:affordanceawareoptions-modelfree}
		\begin{algorithmic}
		    \STATE {\bfseries Require:} Collection of intents $\mathcal{I}$, Environment $E$, Intent completion function $I_{\omega, g}$, Training iterations $mtrain$.
            \STATE {\bfseries Require:} Number of options, Maximum option length $n_{max}$, Attention type. 
            \STATE Initialize parameters $z, \theta$ for $\texttt{IC}^{net}_{z}, \pi_{\omega,\theta}, \beta_{\omega,\theta}$
		    \FOR{$m=1 \dots mtrain$ }
    			\STATE $\mathcal{D}, \mathcal{D}_{\tau} \leftarrow \texttt{DataGeneration}$  using Alg.~\ref{alg:dataset_trajectoryoptions}.%
    			\STATE \textcolor{gray}{// compute gradients over batch}
    			  \STATE Update $\pi_{\omega, \theta}(a| s)$ via PPO towards $I_{\omega, g}$ as reward. ($\mathcal{D}$)
    		      \STATE  Update $\beta_{\omega, \theta}(s')$ with BCE Loss via $I_{\omega, g}$ as target ($\mathcal{D}$)
    		      \STATE Update $\texttt{IC}^{net }_{z}(s, \omega)$ with MSE loss towards Eq.~\ref{eq:affordancetarget} ($\mathcal{D}_{\tau}$)
	      \ENDFOR
		\end{algorithmic}
\end{algorithm}

\textbf{Option Training.} Given a dataset of subgoal-option trajectories with intent completion labels, we compute and propagate the gradients for the respective losses as in Alg.~\ref{alg:affordanceawareoptions-modelfree}. All components of options are parameterized and trained as follows. We use a common neural network backbone and different fully connected layers for options policies, terminations and value estimations. Option policies are trained with proximal policy gradient methods (PPO) ~\citep{schulman2017proximal}), with the objective of  maximizing intent completion. Since we would like the termination parameterization to learn to predict termination condition for all options at the same time, it is trained with standard binary cross entropy (BCE) loss. Finally the $\texttt{IC}^{net}$ is trained using  mean-squared-error (MSE) loss towards the intent completion target, for all partial option trajectories (Eq.~\ref{eq:affordancetarget}). The predictions from the $\texttt{IC}^{net}$ are used to compute the respective attention mechanisms described above. Implementation and training details are provided in App~\ref{sec:appendix-algorithmdetails}.

\section{Empirical Analysis}
\label{experiments}
To investigate the role of attention in decision making with temporal abstraction, we empirically tackle the following questions: \textbf{Q1.} Does hard attention result in improved sample efficiency during online data generation? \textcolor{blue}{Yes} (See Sec.~\ref{sec:sampling}). \textbf{Q2.} Does an HRL agent with affordances (hard attention) perform better in long-horizon sparse reward tasks? \textcolor{blue}{Yes}. In harder tasks, hard attention outperforms soft attention, no attention, and a flat agent with pseudo-reward, potentially due to improved credit assignment. See results on MiniGrid supporting this in Sec.~\ref{sec:longhorizonsparsereward}. \textbf{Q3.} Does hard attention help when increasing the number of choices? \textcolor{blue}{Yes} (See Sec.~\ref{sec:increasingchoices}). \textbf{Q4.} Does our analysis scale to continuous control? \textcolor{blue}{Yes} (See Sec.~\ref{sec:fetchexperiments}). 

\textbf{Motivation for choice of domains.} %
We choose MiniGrid and Fetch domain to allow for 1. specification of intentions, 2. varying complexity in terms of number of options and tasks, 3. isolation and focused investigation of the role of attention, and 4. to demonstrate that our approach scales to both discrete and continuous domains.
 Implementation  details along with the hyperparameters used for all the experiments\footnote{Code for  all  the  experiments  is  available  on \url{https://github.com/andreicnica/hrl_attention}} are provided in App~\ref{sec:appendix-algorithmdetails}.

\subsection{Discrete Domain: MiniGrid}

\paragraph{Environment \& Task Specification.} We use the MiniGrid~\citep{gym_minigrid}, adapted for different configurations of sub-goals. The agent receives a $2D$ fully observable, egocentric, multi-hot encoded view of the map. %
The action space is: turn left, turn right, move forward, pick up. A task is specified by the number of unique objects, is completed when the agent collects all the objects on the map in the correct sequence, upon which the agent receives a positive reward. At the beginning of each episode, a task is sampled at random from a distribution of tasks. There are no common objects across tasks, can differ in type, colour, and disappear from the map once collected. %

\subsubsection{On The Role Of Attention In Data Generation}
\label{sec:sampling}

\begin{wrapfigure}[11]{R}[-0cm]{7cm}
    \begin{minipage}{0.5\textwidth}
        \vspace{-1.0cm}
        \begin{figure}[H]
        \centering
        {\includegraphics[width=0.8\textwidth]{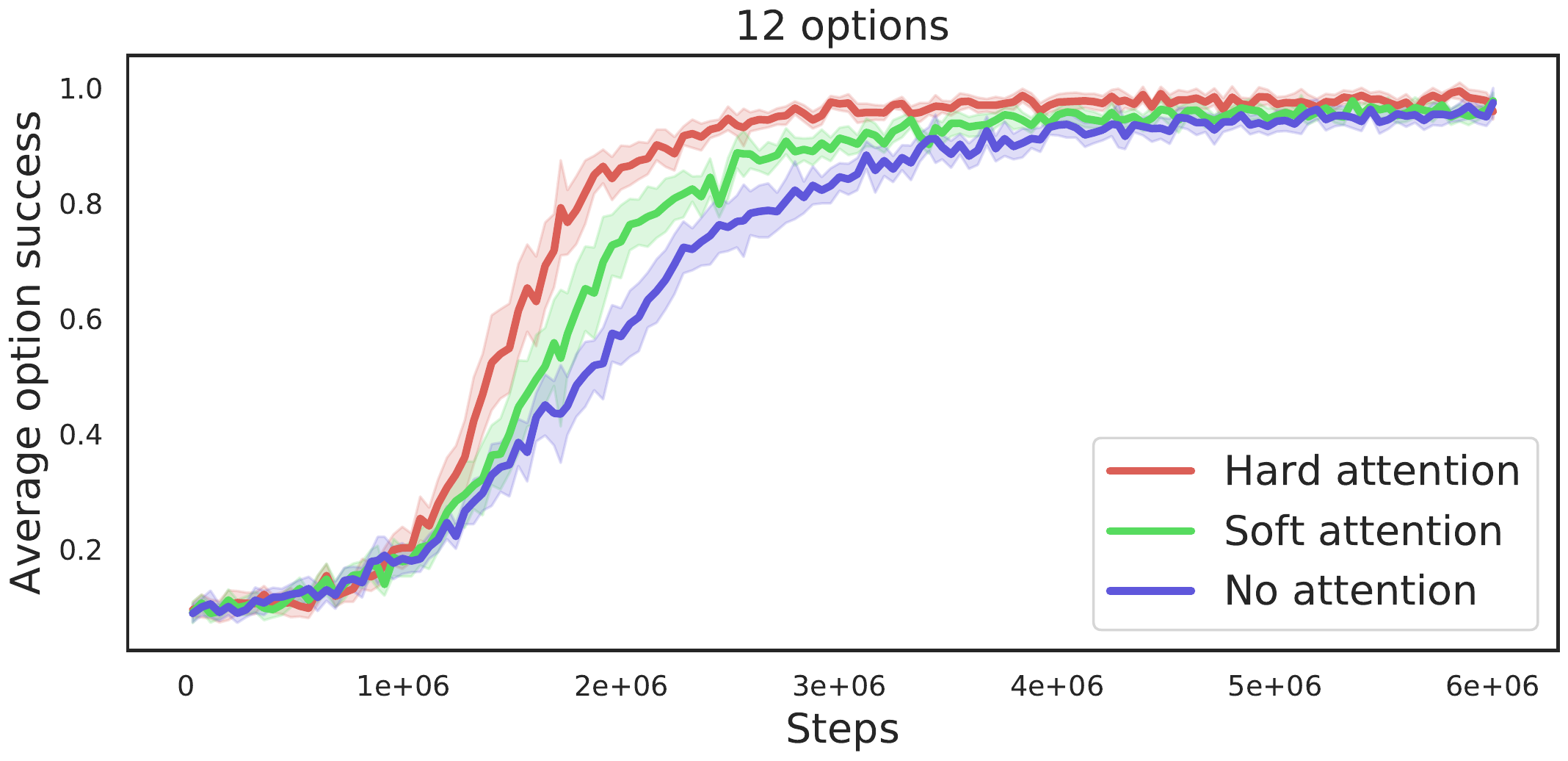}}
        \vskip -0.1in
        \caption{\label{fig:sampling234x3smdpvaluelearning} 
        Option policy evaluation shows that sampling with hard-attention results in faster policies convergence. We plot the average of 10 runs and $95\%$ confidence intervals.}
        \end{figure}
    \end{minipage}
\end{wrapfigure}

\textit{Experimental Setup.} We first evaluate the role of different types of attention while training option policies i.e. during pre-training. We use 4 tasks, each with 3 unique objects to be collected at the beginning of the episode. The agent has to learn 12 options, from which only a maximum of 3 can achieve their subgoal at any given time. A task is sampled randomly, rewarding each option only when it completes its intended subgoal. A new option has to be selected based on the choice of the sampling method.%
 The option policy, termination, and attention mechanism are learned simultaneously. During the training of option policies, we also train the $\texttt{IC}^{net}$, which predicts the discounted intent completion. This poses the challenge of online data generation for learning both the options and attention (see Sec.~\ref{sec:method}). To gain a better understanding of the role of attention during sampling, we compare and highlight differences when data is generated using different attention mechanisms i.e. 1) \textbf{hard-attention}: hard thresholding of the predicted discounted intent, 2) \textbf{soft-attention}: normalized predicted discounted intents and 3) \textbf{no-attention}: sampling uniformly randomly from all choices.

\textit{Results.} For a fair comparison of the quality of the learned options, the option policies are evaluated after each update by comparing their success rate in achieving their corresponding subgoal intent. For evaluation we roll out a set of 64 ground truth affordable options from random initial states. We observe in Fig.~\ref{fig:sampling234x3smdpvaluelearning} that sampling from \textit{hard} attention improves the learned option policies, compared to sampling with \textit{soft} attention or no attention. Leveraging affordances as \textit{hard} attention yields a sample efficient approach to generate training data. $\texttt{IC}^{net}$ learns faster than the option policies, corroborated also by further analysis of its precision during training in App~\ref{sec:appendix-icprecisionwhiletraining}. As a result, knowing which option is strictly affordable narrows the data distribution to fewer but more useful choices. This speeds up the option learning process, as the dataset required by Alg.~\ref{alg:affordanceawareoptions-modelfree} contains more samples of successful option trajectories. Instead, soft-attention dilutes the preferences across all option choices. Despite the success of \textit{hard}-attention in our empirical evaluation, it is intuitive to see that \textit{soft}-attention might be a better choice in some scenarios. See App~\ref{sec:appendix-whenhardattentionmightfail} and ~\ref{sec:appendix-additionalexperiments} for discussion and additional experiments respectively.

\begin{figure*}[t]
    \centering
    {\includegraphics[width=0.77\textwidth]{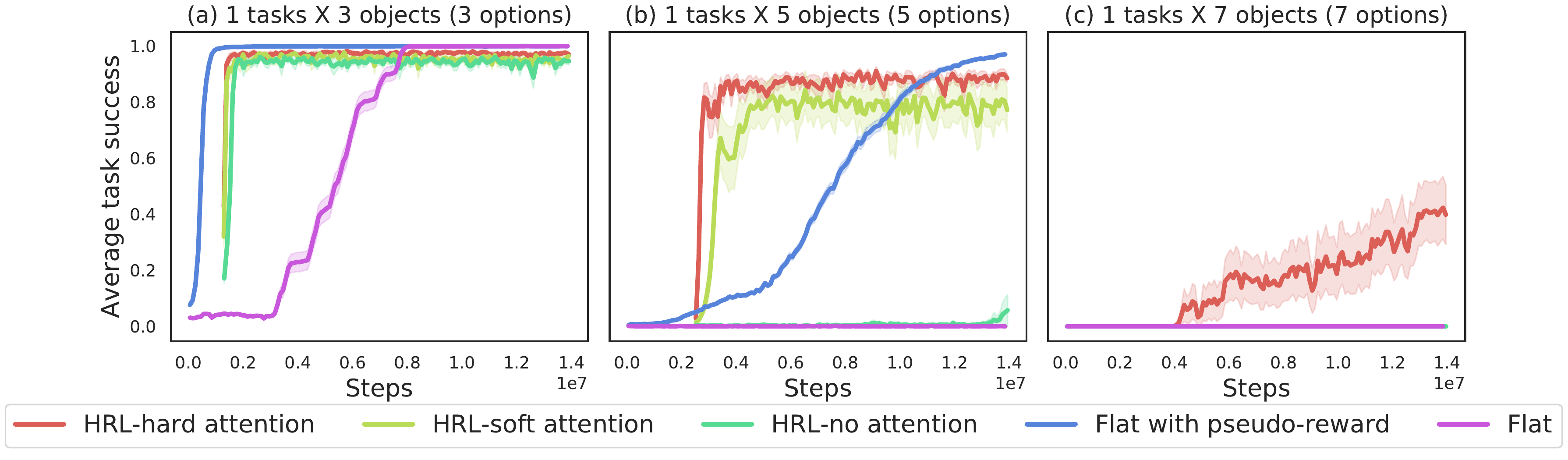}}
    \vskip -0.1in
    \caption{\label{fig:1x357smdpvaluelearning} Long Horizon Sparse Reward Tasks with progressively increasing horizon, via $3$ to $7$ sequential objects. With increasing difficulty, the gap between the hard and soft attention increases. The lines depict the average of 10 runs and $95\%$ confidence intervals.}
        \vskip -0.0in

\end{figure*}

\subsubsection{On The Role Of Attention In Leveraging Previously Acquired Skills}
\label{sec:smdpaluelearning}
Next, we train HRL agents at the SMDP level using PPO algorithm to learn an optimal policy over (pre-trained) options to solve the downstream task. We first learn a set of subgoal-options via Alg.~\ref{alg:affordanceawareoptions-modelfree} in an environment $E$ using different types of attentions. The agent now samples an option based on the learned policy over options ($\pi_\Omega(\omega|s)$) modulated by this attention (Alg.~\ref{alg:attention-modelfree})  as follows:\begin{align}
    \pi_\Omega(\omega|s) \propto f_{att}(s, \omega) \pi_\Omega(\omega|s)
    \label{eq:attentionmodulation}
\end{align}

The sampled option policy ($\pi_\omega(a|s)$) runs until the learned termination ($\beta_\omega(s)$). $\pi_\Omega(\omega|s)$ is learned using the Generalized Advantage Estimator (GAE): 
$\hat{A}^{GAE(\gamma, \lambda)}_{t} = \sum_{l=0}^{T}(\gamma^{m}\lambda)^{l}\delta_{t+1}^{V}$,
where $\delta_{t}^{V} = r_{t} + \gamma^{m}V(s_{t+1}) - V(s_{t})$, $r_t$ is the task-specific reward, $\lambda$ is a hyper-parameter for horizon, $\delta_{t}^{V}$ is the temporal-difference residual, and $\hat{A}^{GAE}$ the advantage estimation. Both $\delta_{t}^{V}$ and $\hat{A}^{GAE}$ are discounted via $\gamma^{m}$ in order to account for the option duration $m$. %

We compare the following methods (See App~\ref{sec:appendix-baselinedetails} for implementation details) 1) A \textbf{flat agent} (PPO) using primitive actions, 2) An \textbf{HRL agent with no attention}, 3) An \textbf{HRL agent with \textit{hard}-attention}, and 4) An \textbf{HRL agent with \textit{soft}-attention}. The HRL agents use pre-trained options with the same attention as in the downstream task and the maximum option pre-training steps are determined based on the option policy performance plateau. We now investigate two settings, namely long horizon sparse reward tasks (Sec.~\ref{sec:longhorizonsparsereward})) and increasing number of option choices (Sec.~\ref{sec:increasingchoices})).

\subsubsection{Long horizon sparse reward tasks}
\label{sec:longhorizonsparsereward}
\textit{Experimental Setup.} The environment is comprised of only one task, but with different numbers of objects to be collected per task (i.e. $3, 5, 7$). The tasks are progressively harder due to the delayed reward with increasing number of objects. %

\textit{Results.} In Fig.~\ref{fig:1x357smdpvaluelearning} we compare the aforementioned methods alongside an additional baseline of the flat agent with a pseudo-reward for whenever the agent collects any of the objects, akin to the subgoal information. We observe that as the total task length is progressively longer, the HRL agent copes better in this more challenging situation as compared to the flat agent (purple) due to faster reasoning at the SMDP level. More importantly, we notice that the HRL agent with \textit{hard}- attention (red) consistently outperforms HRL agent with \textit{soft}- attention due to the agent's ability to 1) strictly restrict its attention to much fewer choices and 2) to cope with imperfect option policies. Our interpretation of this result is that limiting an agent's attention to fewer but meaningful choices potentially results in improved credit assignment particularly in such long-horizon sparse reward tasks. While the flat agent with pseudo-reward outperforms the options agent in easy tasks (Fig~\ref{fig:1x357smdpvaluelearning}a.), in more challenging longer horizon tasks (Fig~\ref{fig:1x357smdpvaluelearning}c.), it is unable to learn even with significantly high number of training steps. In this harder task the  HRL agent with \textit{hard}- attention is the only one able to learn.

\subsubsection{Increasing number of choices}
\label{sec:increasingchoices}
\begin{figure*}[]
    \centering
    \includegraphics[width=0.69\textwidth]{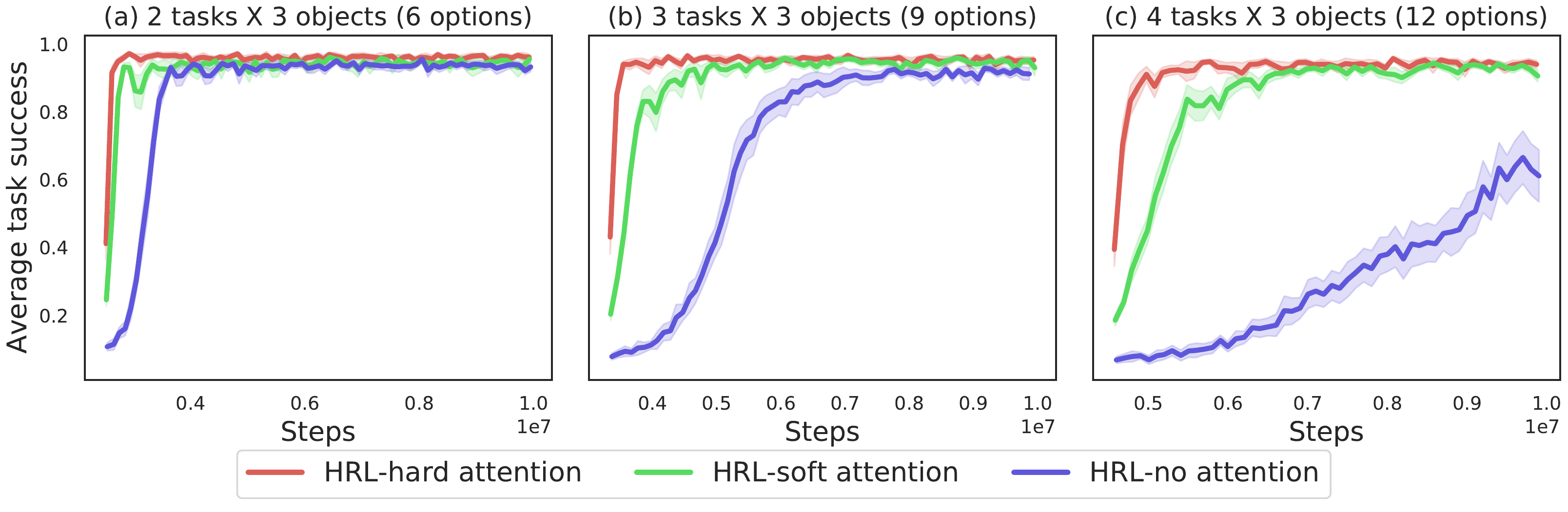}
    \vskip -0.1in
    \caption{\label{fig:234x3smdpvaluelearning} SMDP value learning with increasing number of choices (using pre-trained options). The gap between hard and soft attention increases as more choices are to be considered demonstrating the paradox of choice. We plot the average of 10 runs with $95\%$ confidence intervals.}
    \vskip -0.0in
\end{figure*}
\textit{Experimental Setup.} Next, we investigate the role of attention with growing number of choices, by varying only the number of tasks. The agent is tasked to collect a fixed number of objects ($3$ in this case) per task in the correct order. Leveraging the pre-trained options, we learn a policy over options in order to solve the task (as detailed in Sec.~\ref{sec:smdpaluelearning}). 

\textit{Results.} Fig.\ref{fig:234x3smdpvaluelearning} depicts
the average task success for increasing number of choices from left to right, namely $6, 9,$ and $12$ options. We observe that as the decision maker has more choices to consider, restricting the agent's action possibilities strictly (such as via \textit{hard}-attention) results in better performance highlighting the paradox of choice. When smaller number of choices are to be considered, both \textit{soft} and \textit{hard}-attention perform almost the same while the agent without any attention is slow in learning. We anticipate with growing number of choices, knowing what is affordable as a \textit{hard}-attention criteria enables agents to be much more sample efficient than agents that dilute attention over all choices (green) and agents that do not attend selectively (purple).

\begin{wrapfigure}[10]{r}[-0mm]{7cm}
    \begin{minipage}{0.5\textwidth}
        \vspace{-1.78cm}
        \begin{figure}[H]
            \centering
            \subfloat[$\texttt{IC}^{net}$\label{fig:pickupthekey_intentprediction}]{\includegraphics[width=0.28\textwidth]{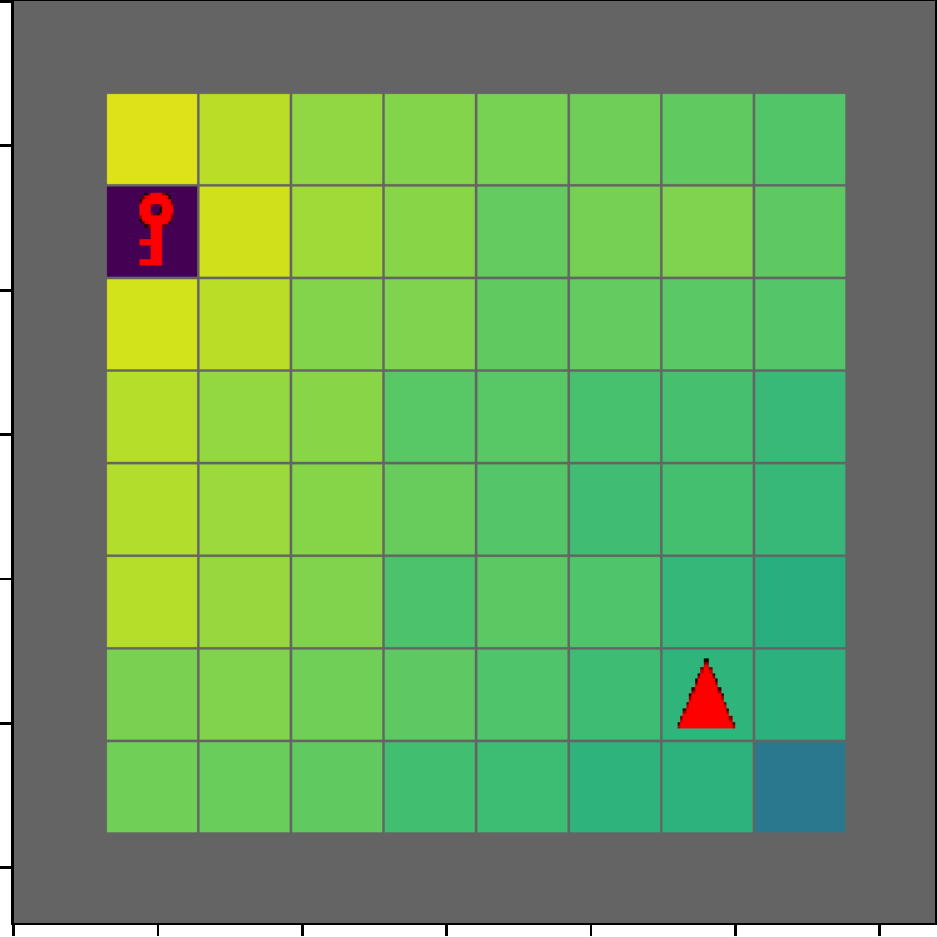}} \hspace{4mm}
            \subfloat[Affordances\label{fig:openthedoor_hardthreshk10}]{\includegraphics[width=0.28\textwidth]{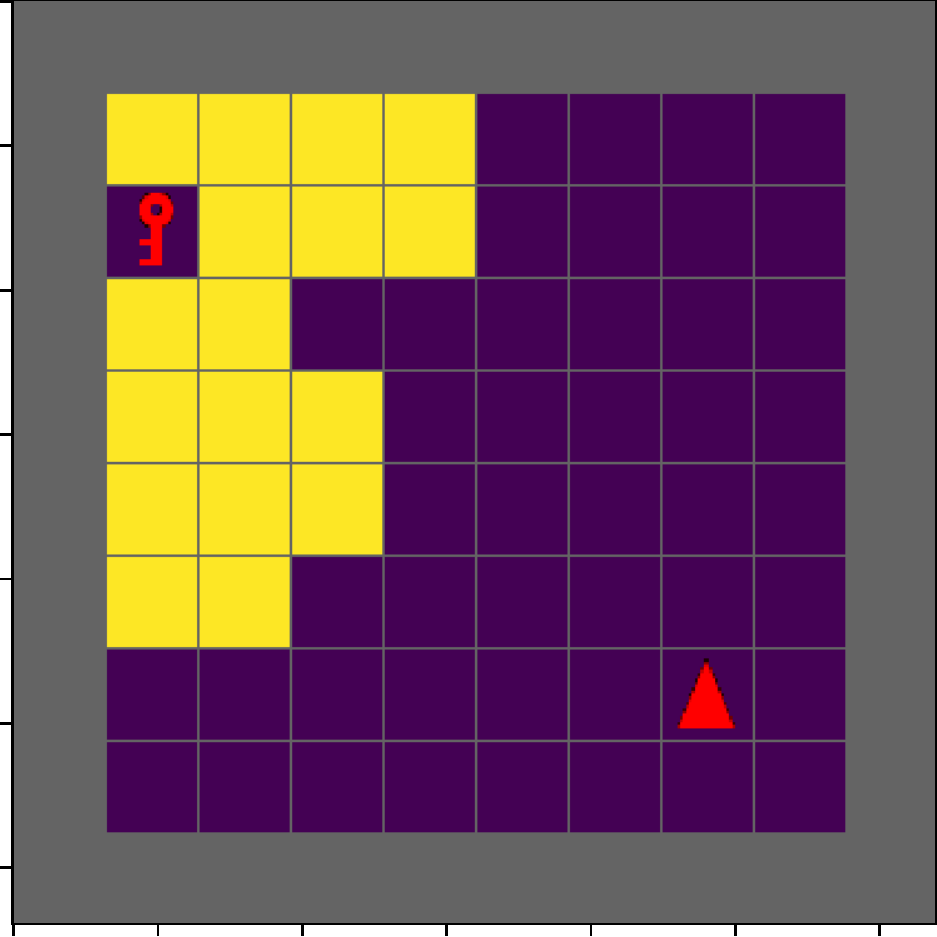}} \hspace{4mm}
            \includegraphics[width=0.065\textwidth]{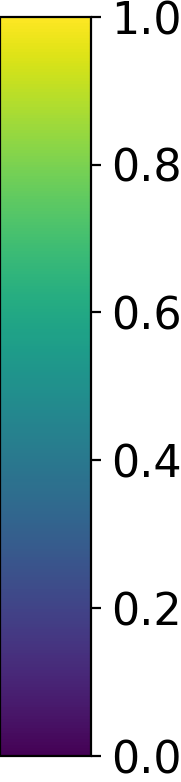}
            \vskip -0.1in
            \caption{Qualitative Analysis shows (a) predictions from $\texttt{IC}^{net}$ for \textit{pick-up-the-key}. (b) A hard thresholding on $\texttt{IC}^{net}$ predictions generates option affordances (threshold $\gamma^{10}_{\texttt{IC}^{net}}$).
    \label{fig:qualitativeanalysis}}
            \vskip -0.1in
        \end{figure}
     \end{minipage}
\end{wrapfigure}
\subsubsection{Qualitative Analysis}

We analyze the predicted discounted intent completion target (Eq.~\ref{eq:affordancetarget}) and affordances qualitatively via a heatmap in MiniGrid in Fig.~\ref{fig:qualitativeanalysis}. We observe that learned subgoal-intent completion is peaked in states around the subgoal such as near the key (Fig.~\ref{fig:pickupthekey_intentprediction}). Higher values are denoted in brighter yellow while the lower ones in darker colors. See additional examples in App~\ref{sec:appendix-attentionAnalysis}.
To obtain option affordances, we apply a hard threshold on the $\texttt{IC}$ network predictions. Here we plot the affordances for a threshold of $10$ steps, restricting the option to be affordable for intent completion values $\geq \gamma_{IC}^{10}$.%

\subsection{Continuous Control: Fetch Domain}
\label{sec:fetchexperiments}
\begin{wrapfigure}[10]{R}[-0cm]{7cm}
    \begin{minipage}{0.5\textwidth}
        \vspace{-1.3cm}
        \begin{figure}[H]
            \centering
            \includegraphics[width=0.8\textwidth]{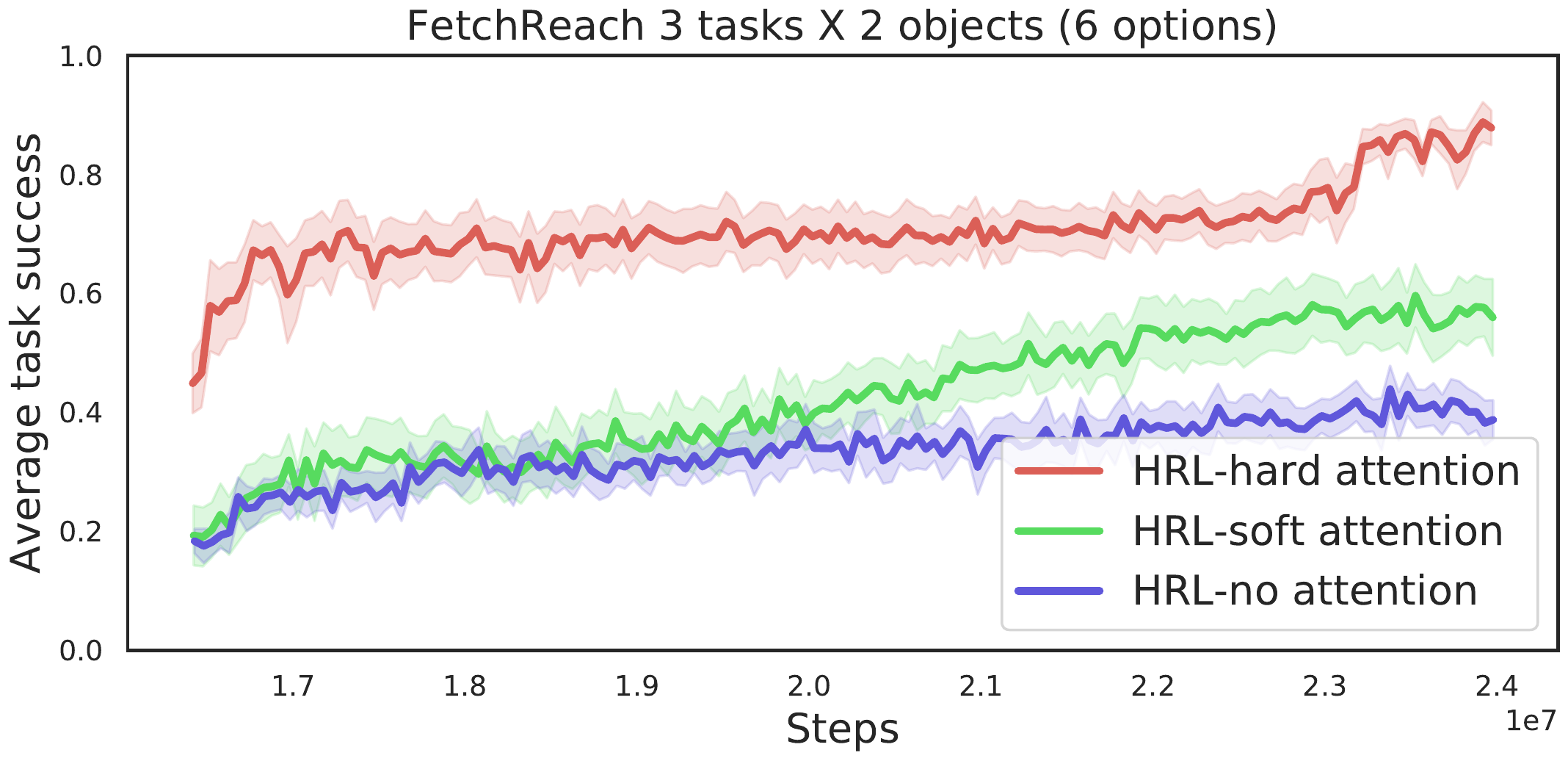}
            \vskip -0.1in
            \caption{\label{fig:2x3fetch} SMDP value learning in Fetch Reach. We plot the average of 10 runs with $95\%$ confidence intervals.}
            \vskip -0.1in
        \end{figure}
    \end{minipage}
\end{wrapfigure}

\textit{Experimental Setup.} To demonstrate choice attention in HRL scale for continuous control robotics environment, we consider tasks with multiple goals per task in an adaptation of the FetchReach-v0 domain ~\citep{schulman2017proximal}. The agent must learn to move a robotic gripper close to a set of target positions in the correct order. All goal positions are represented in the input observation by either a valid coordinate from the state space or a null coordinate for unavailable targets. A sparse reward is used for both option policy training (intent completion) and downstream task (See Sec.~\ref{sec:smdpaluelearning}). See Appendix~\ref{sec:appendix-fetchdetails} for implementation details.

\textit{Results.} Fig.\ref{fig:2x3fetch} shows that affordance aware subgoal options have an advantage in learning %
even when considering $6$ options in continuous space. Although the option policies are limited from pre-training (lower maximum downstream reward possible), affordance aware SMDP learning can still show a major advantage in performance in significantly lesser time. See App~\ref{sec:appendix-fetchadditionalexperiments} for option policy and HRL training evaluation for different number of goals.

\section{Related Work}
Attention is a widely studied concept in the vision and language community. For e.g., when considering images, \textit{soft}-attention places learned multiplicative weights over \emph{all} patches in the input image, ~\cite{bahdanau2014neural}; soft-attention has access to the entire image. On the other hand, \textit{hard}-attention only selects one patch of the image to attend to~\citep{luong2015effective, xu2015show}. This already established distinction can be juxtaposed with the characterization of attention that we use in our work in terms of action choices. Specifically, \textit{hard}-attention via affordances only allows certain option choices and strictly limits the agent's focus to parts of the option space. \textit{Soft}-attention, on the other hand, places preferences over \textit{all} actions.

Learning options has been extensively explored in the literature~\citep{parr1998reinforcement,thrun1995finding,dayan1993feudal,dietterich2000hierarchical,stolle2002learning, mcgovern2001automatic}. Our work can be viewed as leading to options with initiations sets~\citep{sutton1999between}. \citet{khetarpal2020i} proposed end-to-end gradient based discovery to learn parameterized interest functions, a generalization of initiation sets, by optimizing a task specific reward signal. In contrast, we consider learning options with given subgoals, akin to using pseudo-rewards~\citep{kulkarni2016hierarchical}. 

Our work is related to \textit{hierarchical affordance competition}~\citep{pezzulo2016navigating}, where interactive decision making entails a competition between representations of available actions (affordances) biased by the desirability of their predicted outcomes (intents). Here, we consider that we are given desirable outcomes of an \textit{intentional action} via the specification of subgoals. The affordances, as a \textit{hard}-attention mechanism, facilitate reasoning across different time scales, by predicting the expected future outcomes of options using the discounted intent completion. In RL, the closest to our work is \citet{xu2020deep}; a method that incrementally builds environment models around the affordances of parameterized motor skills. Unlike their work, we propose a model-free online algorithm. %
We also assume that subgoal information is known apriori and utilized as intents. More recent, concurrent work \citep{khetarpal2021temporally} considers a model-based approach in building partial option models based on affordances and is complementary to our work.

\section{Discussion}
\label{sec:discussion}
We introduced a notion of affordances for goal-based options, defined through a \textit{hard}-attention mechanism, and presented an approach for learning both option policies and the associated affordances, when subgoals for the options are given. Our empirical results demonstrate that this approach can lead to fewer but more meaningful choices, compared to soft preferences over all option choices. Furthermore we show that using attention mechanisms for HRL contributes to faster learning and higher quality training data in terms of rewarding trajectories. We have focused on two types of domains, discrete  (MiniGrid) and continuous (Fetch). While relatively small, they allow us to do careful quantitative and qualitative analysis, to control the number and difficulty of the tasks. %
We expect our approach to be especially useful in tasks where intent achievement can be easily verified, such as robotics or dialog management tasks. 

In our experiments, we used subgoal-based options because this allows for setting termination conditions, and therefore the intents, to be easily defined, but our approach is general, and could be used together with other types of termination.
We also assumed prior knowledge of the subgoals. This assumption, could be relaxed through a two-step iterative algorithm, 1) subgoal discovery, e.g. using an information theoretic objective, such as in~\citet{harutyunyan2019termination}, 2) iterative learning of affordance-aware subgoal options. However, note that learning intents, affordances and options simultaneously could make it harder to study hard vs soft attention. The choice of intents/subgoals presented provides ground truth for the hard attention, which facilitates the evaluation and brings clarity in reporting. We emphasize that our analysis is complementary to {\em any} method providing intents.

\begin{ack}
The authors would like to thank Emmanuel Bengio, Joelle Pineau for detailed feedback, Zafarali Ahmed for valuable comments on a draft of this paper. A special thank you to Emmanuel Bengio for helping us edit the paper. The authors also thank the anonymous ICML reviewers and AAAI reviewers.
\end{ack}

\bibliography{references}
\bibliographystyle{apalike}

\appendix

\onecolumn
\renewcommand\thefigure{\thesection\arabic{figure}}    
\setcounter{section}{0}
\setcounter{figure}{0}
\section{Appendix}
\label{sec:appendix}

\subsection{Discussion: Limitations of \textit{Hard-}Attention for Online Data Generation}
\label{sec:appendix-whenhardattentionmightfail}
In our empirical analysis, while we observed that \textit{hard}-attention results in a more sample efficient sampling approach during online data generation, we here discuss potential drawbacks of using \textit{hard}-attention for online data generation. A potential issue with sampling from \textit{hard}-attention (affordances) can be a deficient training dataset due to multiple reasons such as (i) the initial stages of training has randomly initialized affordances resulting in poor choices of option selection for training from those states, further highlighting the chicken-and-egg problem, (ii) the dependence on the hard threshold can often be restrictive for dataset generation especially for affordance classification which requires both positive and negative samples, and (iii) when multiple options are affordable in a given state, the initial stages of learning bias the network towards one or the other choice resulting in further imbalance in the dataset. To cope with these issues, it might be beneficial to use \textit{soft}-attention in the exploratory phase of an RL agent. 

\subsection{Algorithm Implementation Details}
\label{sec:appendix-algorithmdetails}
\subsubsection{Algorithm Details}

We consider $I_{\omega, g}$ is subgoal intent specification known apriori. For practical purposes, intent completion requires an option trajectory as input, and returns a binary value indicating if the intent was completed or not. Speficially, for the object world in minigrid, the indicator function i.e. $\mathbbm{1}_{g_{w}}(\tau_{i, s_{1}:s_{t}})$ returns $1$ if the corresponding subgoal object is collected, otherwise $0$.

\subsubsection{Architecture \& Training Details}
\label{sec:appendix-trainingdetails}
\textbf{Common details} for pre-training of options and SMDP value learning are as follows: We define the following neural network common architecture design choices. As a backbone for extracting features from the input, we use $3$ convolution layers with kernels $[32, 32, 64]$ of size $3$ and strides $[2, 1, 1]$ with $1$ fully connected layer. All fully-connected layers have dimensionality 256. All layers use RELU non-linearities \citep{nair2010rectified}. All hyper parameters are provided in Table~\ref{table:hyper_params}.

\textbf{Option Training.} We use the clipped version of the PPO~\citep{schwartz2004paradox} for training options. The network is composed of a architecture including the backbone (as described above) and a fully-connected 2-layer head for each option policy, each termination module and the value estimator. A separate network with same architecture as backbone and fully-connected layer is used for predicting the discounted intent completion values for a state and option id. The input of the network is represented by the state concatenated with a one hot encoded representation of the option id. 

\textbf{SMDP Value Learning.} To train the policy over options we train a neural network composed of the backbone architecture and two fully-connected 2-layer heads (actor and critic) to choose from the available number of options given the state of the environment. We use PPO. We use the SMDP updates as described in Sec.~\ref{sec:smdpaluelearning} of the main paper. 

\begin{table}[h!]
\centering
\begin{tabular}{|l|c|}
\hline
Parameter Name & Value \\
\hline
\textbf{PPO algorithm (Option pre-training / HRL Learning)} & \\
Frame rendering & $21 \times 15 \times 15$ \\
Number of parallel environments & $16$ \\
Rollout length & $128$ / $64$ \\
Optimizer & RMSProp \\
Learning Rate & $0.0003$ \\
Optimization Epochs & $4$ \\
batch size & $256$ \\
discount & $0.99$ \\
GAE lambda & $0.95$ \\
entropy coefficient & $0.01$ / $0.0$ \\
value loss coefficient & $0.5$ \\
termination loss coefficient & $0.5$ \\
clip eps & $0.2$ \\
max grad norm & $0.5$ \\
Termination Threshold (HRL) & nan / $0.5$ \\
Maximum trajectory length & $50$ \\
\hline
\textbf{Intent completion network training} & \\
Frame rendering & ($\text{num}_{\text{options}}$ $ + 21) \times 15 \times 15 $ \\
Optimizer & RMSProp \\
Learning Rate & $0.0003$ \\
Batch Size & $128$ \\
Affordance (hard-attention) Threshold $k$ & $90$ \\
Discount Factor $\gamma_{IC}$  & $0.98$ \\
Normalized $IC_{target}$ values range & $[-1.5, 1.5]$ \\
\hline

\end{tabular}\\
\caption{\label{table:hyper_params} A complete overview of used hyper parameters.}
\end{table}

\subsubsection{Minigrid Experiments: Implementation Details}
We now provide details of the Minigrid experiments. We consider a room size of $10 \times 10$ with an egocentric multi-hot encoded observation of dimensionality $15 \times 15 \times 21$ (where $21$ is comprised of $12$ object types, $6$ color types, and $3$ states.) %

\subsubsection{Baselines Details}
\label{sec:appendix-baselinedetails}
\begin{enumerate}
\setlength\itemsep{0.0001em}
    \item A \textbf{flat agent} (PPO) using primitive actions with comparable amount of experience. For a fair comparison with the primitive agents, all plots show the HRL agents off set by the amount of experience needed for training the options and synced at the number of MDP steps. 
    \item An \textbf{HRL agent with no attention}: it is assumed that all options apply everywhere, and therefore the an option is directly sampled from the the policy over options without any modulation.
    \item An \textbf{HRL agent with \textit{hard}-attention}: an option is sampled from the policy over options which is now being modulated with $f_{att}=\texttt{hard}$ generated using Alg~\ref{alg:affordanceawareoptions-modelfree}. 
    \item An \textbf{HRL agent with \textit{soft}-attention}: an option is sampled from the policy over options which is now being modulated with $f_{att}=\texttt{soft}$ generated using Alg~\ref{alg:affordanceawareoptions-modelfree}.
\end{enumerate}

\subsubsection{FetchReach Experiments: Implementation Details}
\label{sec:appendix-fetchdetails}

In this section, we provide complete details of the Fetch experiments (Sec.~\ref{sec:fetchexperiments}). We represent the same problem setup, that of multiple tasks with multiple goals per task, in a robotic continuous space environment. We choose to adapt the FetchReach-v0 environment from ~\cite{plappert2018multi} in order to train an agent to move a robotic gripper close to a set of target positions in the correct order. We represent all the multiple goal positions in the input space by 3D coordinates, sampled around the gripper starting position. We adapt the original Fetch environment to account for the correct initialization of the goal positions. Specifically, we ensure that all target positions are within the required distance of each other such that no two goals can be collected at the same step. If a goal is not available (either it has already been collected or it is not part of the sampled task) it is represented by the same invalid coordinate (a coordinate which is outside the valid state space, $0$ for our configuration). We use a sparse reward for both policy training and the downstream task.

\paragraph{Adaptations in Architecture \& Training for Fetch.}

For Fetch experiments, we keep the architecture and training same as Appendix Sec.~\ref{sec:appendix-trainingdetails} with the following changes. The backbone for extracting features is 3 layer MLP (size 768) with SELU activation functions. The number of parallel environments is 32. The optimizer is set to be Adam, with a learning rate of 0.0001, optimization batch size of 1024, optimization epoch size of 10, clip eps parameter value to be 0.05, and entropy coefficient to be 0.

\subsubsection{Compute and resources}
\label{sec:appendix-compute}

Experiments from the MiniGrid environment run at approximately $400$ FPS and require approximately 16 hours (including option pre-training and  HRL agent training). The FetchReach environment runs at ~$500$ FPS and one seed experiment requires 24 hours. To ensure reproducibility, we will release more details on compute upon completion of the review process. %

\subsection{Analysis on Choice of Threshold for \textit{Hard} Attention}
\label{sec:appendix-attentionAnalysis}

\subsubsection{Chosen Threshold Analysis for Minigrid}
\begin{figure*}[h]
    \centering
    {\includegraphics[width=0.4\textwidth]{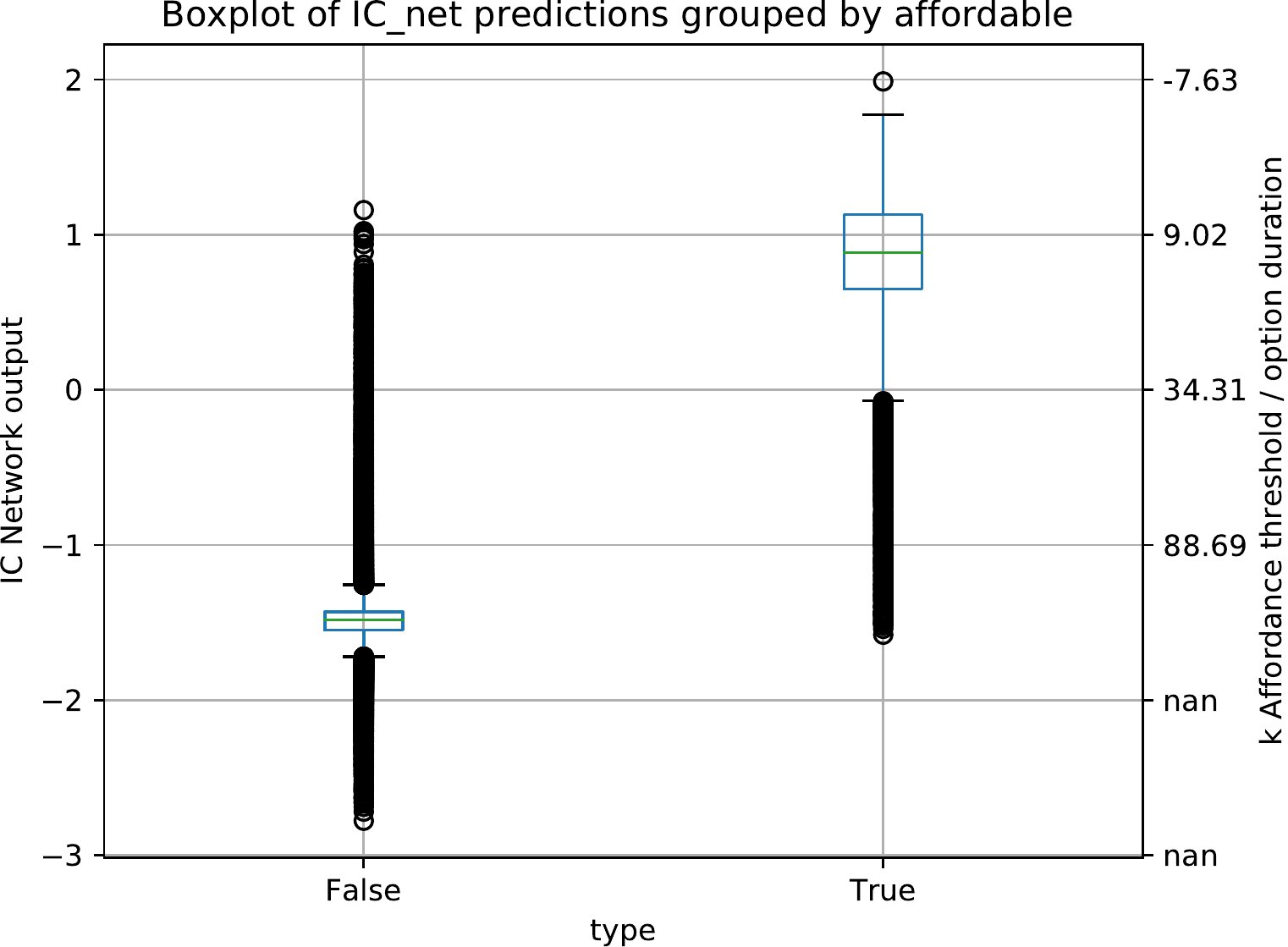}}
    \vskip -0.1in
    \caption{\label{figapp:affboxplot}{$\texttt{IC}_{net}$ Normalized Predicted values.} We observe that our choice of the hard-threshold of $k=90$ separates the two distributions of affordable (True) and non-affordable (False) subgoals with high accuracy.}
\end{figure*}

We now analyze our choice of the threshold for computing \textit{hard}-attention. In particular, we consider the Minigrid environment with $4$ tasks and $3$ objects per task as described before. From the entire state-space of this domain, we consider $10$k randomly sampled observations and all $12$ available options. In Fig.~\ref{figapp:affboxplot} we plot the normalized values predicted by the $IC_{net}$ for these observations. 
 
The box plot values are grouped based on whether the corresponding subgoal-intent being analysed was available in the state or not (affordable or not). For subgoals which are not affordable the $\texttt{IC}_{target}=0$, which is then normalized to the range $[-1.5, 1.5]$, whereas for the subgoals which are affordable, the discounted target is $\gamma_{\texttt{IC}}^{\texttt{optionduration}}$, mapping to the same range. We observe in Fig.~\ref{figapp:affboxplot} that most values for the non-affordable subgoals are at the minimum normalised $\texttt{IC}_{target}$ value as desired. For the affordable subgoals, the chosen affordance threshold of $k=90$ separates the two distributions with high accuracy (accuracy ~$0.98$, precision ~$0.93$, recall ~$0.99$), with the exceptions of some outliers (black dots). We note that most options take the average duration of ~$10$ steps (as depicted here in the right y-axis of the plot) to complete the subgoal intent. To demonstrate this analysis, we trained the network with sampling strategy as \textit{soft}-attention.

Next, we measure the accuracy of affordance prediction made by $IC_{net}$ using the chosen threshold of $k=90$. Speficially, we report the accuracy, precision and recall based on the ground truth values of affordable subgoal intents. We evaluate each network used in all the HRL-hard attention agents from Fig.~\ref{fig:1x357smdpvaluelearning} and Fig.~\ref{fig:234x3smdpvaluelearning} on $10$k randomly sampled observations from the corresponding environments. We observe in Table~\ref{table:affordanceaccuracy} a higher prediction accuracy in simpler environments which gradually decreases with increased complexity of tasks and number of objects per task. We note that the recall score decreases more drastically as the number of objects per task increases and the precision decreases in general with the number of objects. 

\begin{table}[h!]
\centering\renewcommand\cellalign{lc}
\setcellgapes{3pt}\makegapedcells

\begin{tabular}{|l|c|c|c|c|}
\hline
\makecell{\textbf{Environment} \\ (num tasks $\times$ \\ objects per task)} & \makecell{Checkpoint \\ Frames (Mil)} & \makecell{\textbf{Accuracy \%}\\(std dev)} & \makecell{\textbf{Precision \%}\\ (std dev)} & \makecell{\textbf{Recall \%}\\ (std dev)} \\
\hline
1 $\times$ 3 & $\sim 1.24$ & $98.933$ ($0.015$) &  $98.7231$ ($0.022$) & $99.7093$ ($0.002$) \\
\hline
1 $\times$ 5 & $\sim 2.45$ & $97.6874$ ($0.020$) &  $97.254$ ($0.030$) & $98.9837$ ($0.006$) \\
\hline
1 $\times$ 7 & $\sim 3.68$ & $96.3231$ ($0.028$) &  $97.4448$ ($0.036$) & $96.1152$ ($0.016$) \\
\hline
2 $\times$ 3 & $\sim 2.45$ & $98.7708$ ($0.014$) &  $97.1169$ ($0.037$) & $99.3463$ ($0.004$) \\
\hline
3 $\times$ 3 & $\sim 3.27$ & $97.8076$ ($0.023$) &  $92.0896$ ($0.077$) & $99.313$ ($0.007$) \\
\hline
4 $\times$ 3 & $\sim 4.50$ & $95.6185$ ($0.051$) &  $83.0340$ ($0.156$) & $98.6876$ ($0.005$) \\
\hline

\end{tabular}\\
\caption{\label{table:affordanceaccuracy} A complete overview of the quality of predicted affordances for all environments used in the article and their corresponding checkpoint used in the HRL-hard attention agents. The standard deviation is calculated over the 10 random seeds experiments for each HRL-agent.}
\end{table}

\subsubsection{Sensitivity analysis to different thresholds ($k$)}

\begin{figure*}[h]
    \centering
    \subfloat[Pre-training \label{fig:pretraining}]{\includegraphics[width=0.41\textwidth]{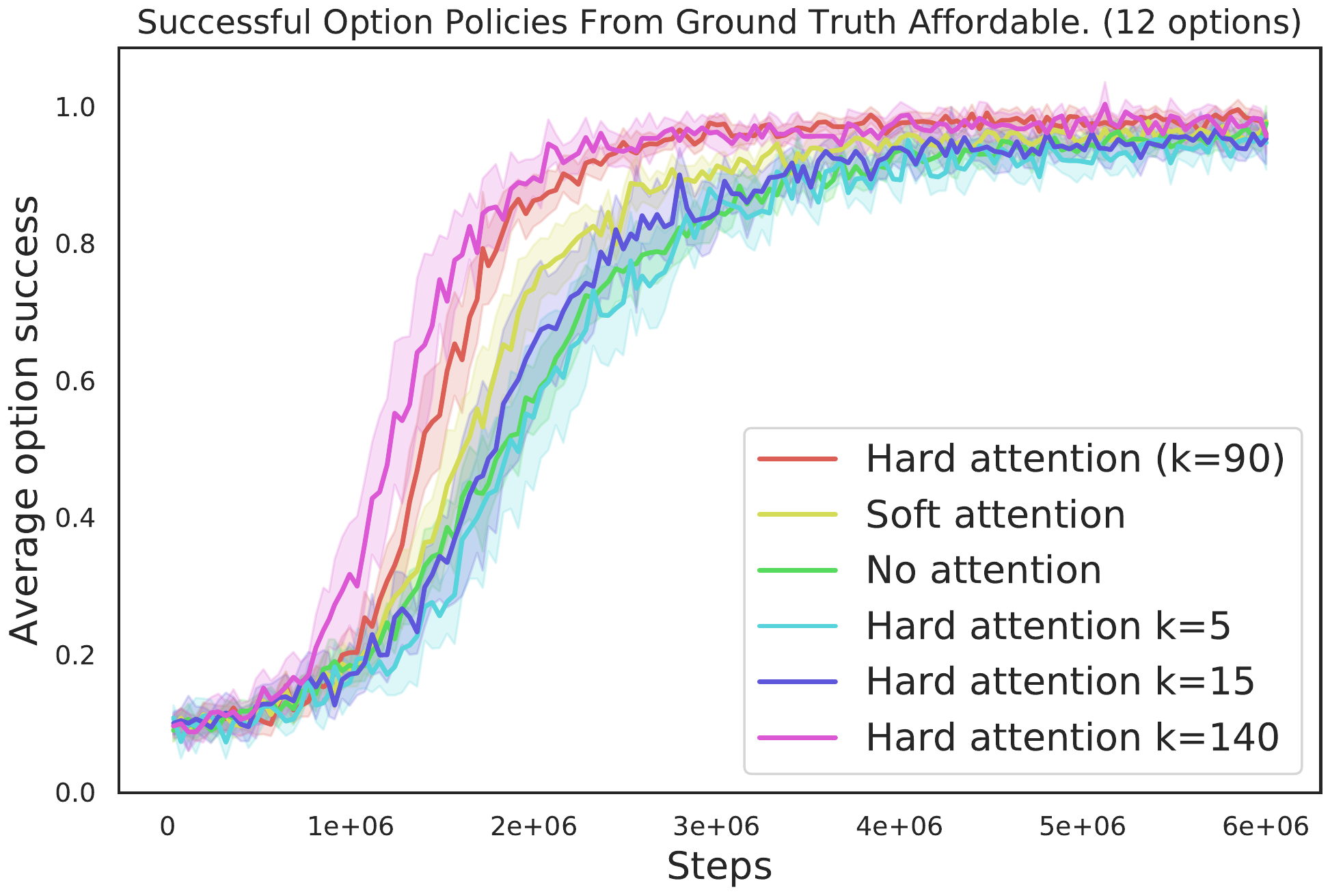}} \hspace{2mm}
    \subfloat[Downstream  Tasks\label{fig:downstream}]{\includegraphics[width=0.4\textwidth]{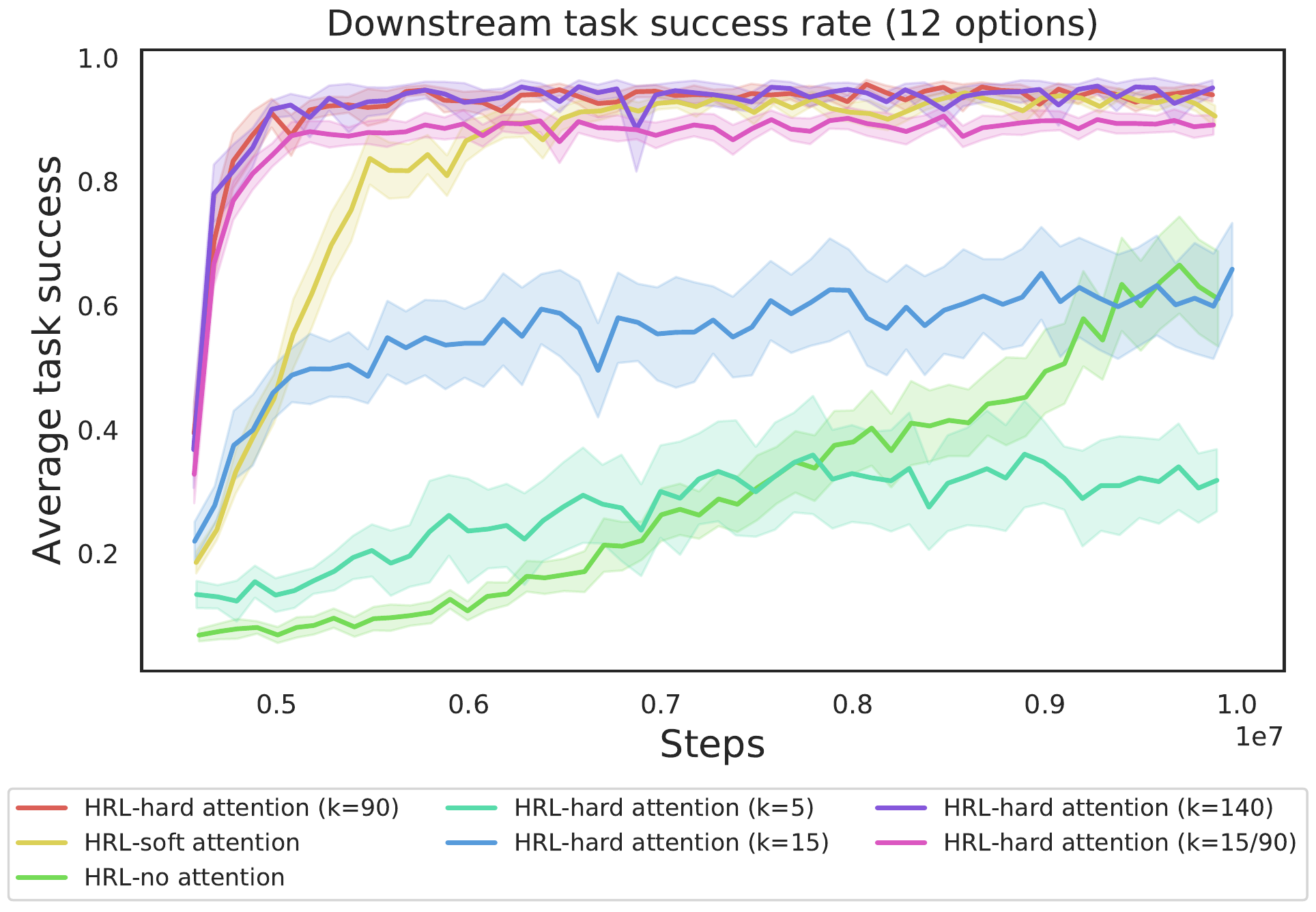}}
    \caption{\label{fig:varyingkanalysis} Impact of varying k during (a) online data generation - learning both option policies and affordances and (b) SMDP value learning for downstream tasks.  The lines depict the average of 10 runs and $95\%$ confidence intervals.}
\end{figure*}

Fig.~\ref{fig:varyingkanalysis} depicts additional results to sensitivity to threshold parameter in both pretraining options and HRL agents. We investigate the sensitivity to varying values of $k\in\{(5, 15, 90, 140\}$. The results corroborate the intuition that lower values of $k$ (e.g. 5, 15) lead to fewer affordable options, including no affordable options for states farther in time (due to our design of affordances via discounted intent), resulting in slower convergence of the learned option policies during pretraining (Fig.~\ref{fig:pretraining}). Larger values of $k$ (e.g 90, 140) result in more options being affordable, which facilitates generating option trajectories with better coverage of the state space,  improving performance in downstream tasks (Fig.~\ref{fig:downstream}). %

Note that for options trained with restrictive hard-attention during learning (e.g. $k=15$), we can recover performance in downstream tasks with a better  threshold ($k=90$). Note that the HRL agent with the combination of $k=15/90$ (Fig.~\ref{fig:downstream}- magenta) still outperforms the HRL agent with soft-attention (green).

\subsubsection{Additional Qualitative Results}
\begin{figure}[h]
\begin{tabular}{lll}
\centerline{$\texttt{IC}^{net}_{z}(s, \omega)$} \vspace{0.1cm}\\
\hspace{-0.2cm}\raisebox{-.5\height}{\includegraphics[scale=0.23]{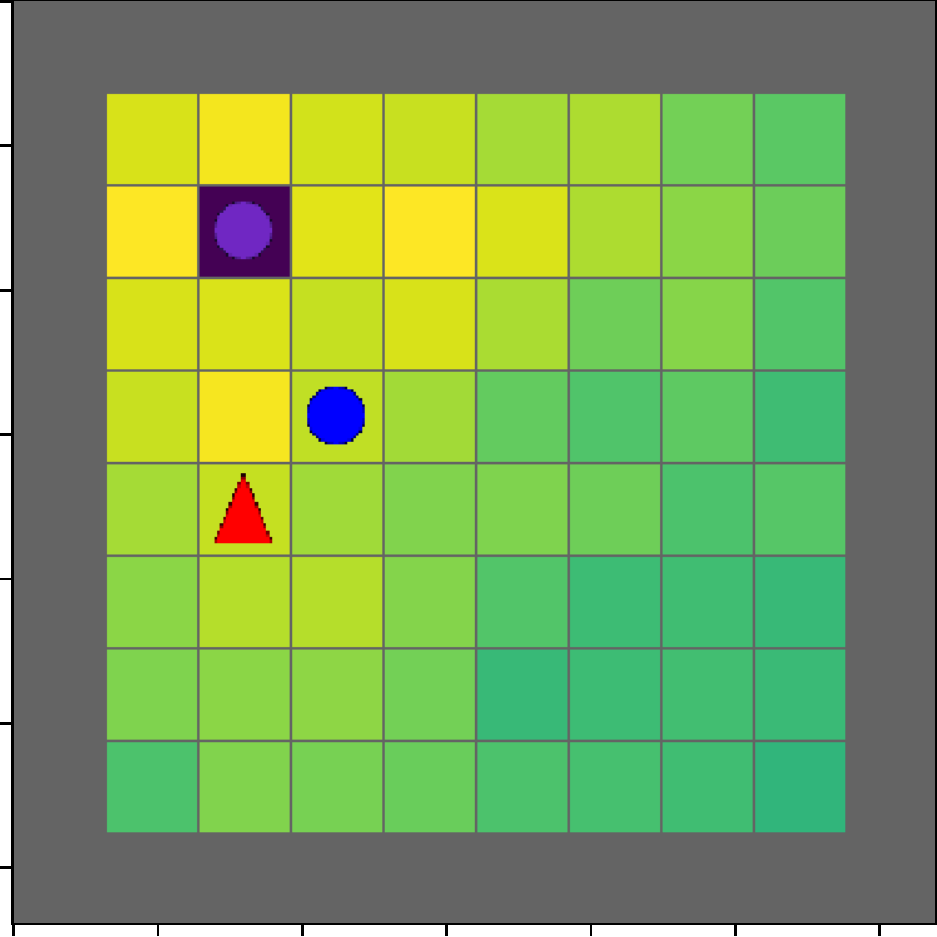}} 
\raisebox{-.5\height}{\includegraphics[scale=0.23]{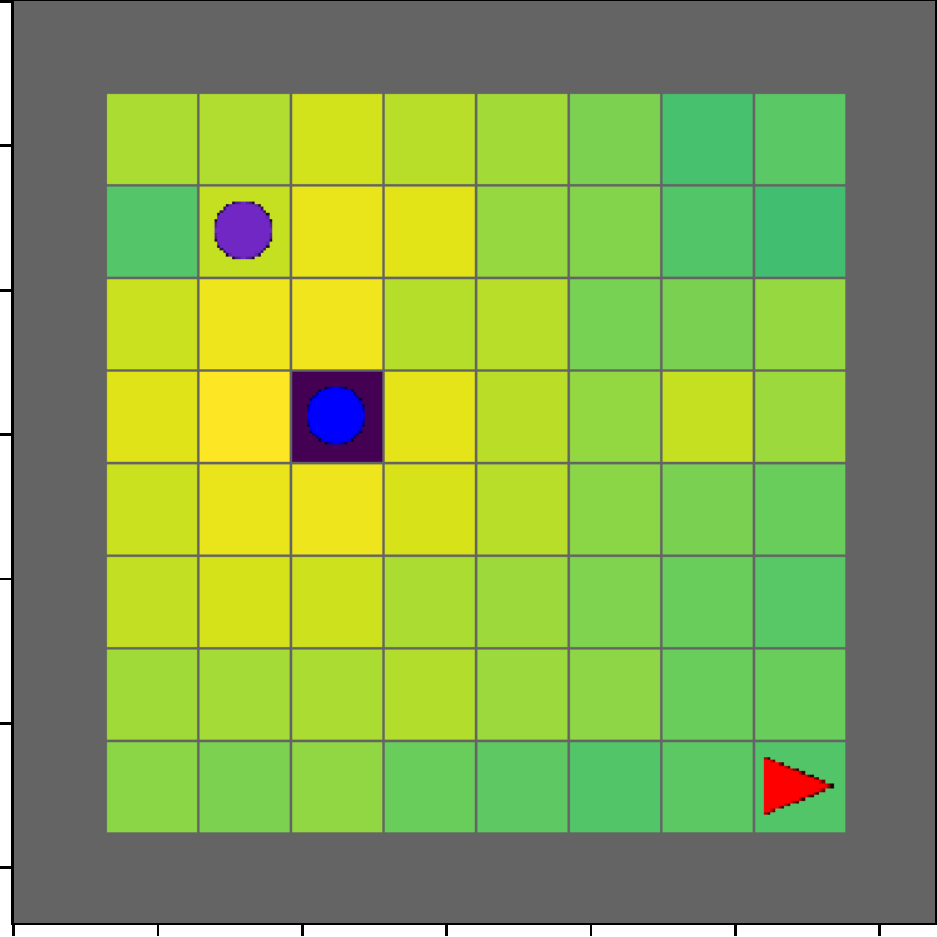}} 
\raisebox{-.5\height}{\includegraphics[scale=0.23]{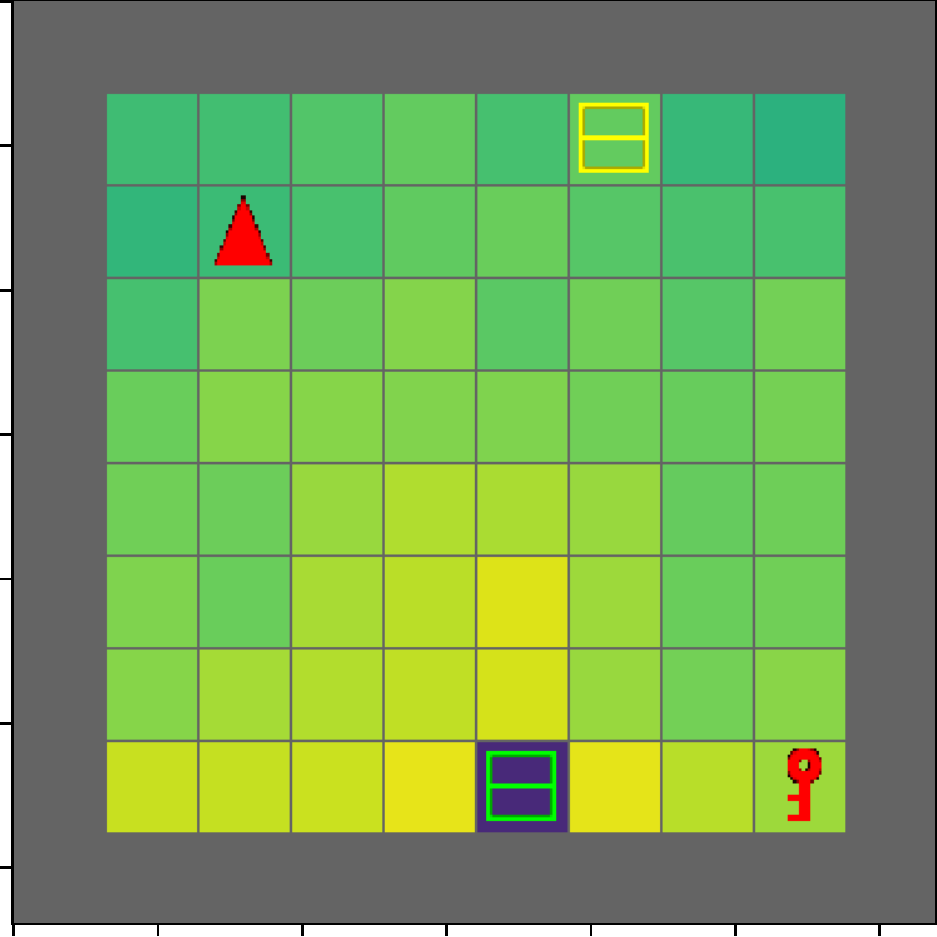}}
\raisebox{-.5\height}{\includegraphics[scale=0.23]{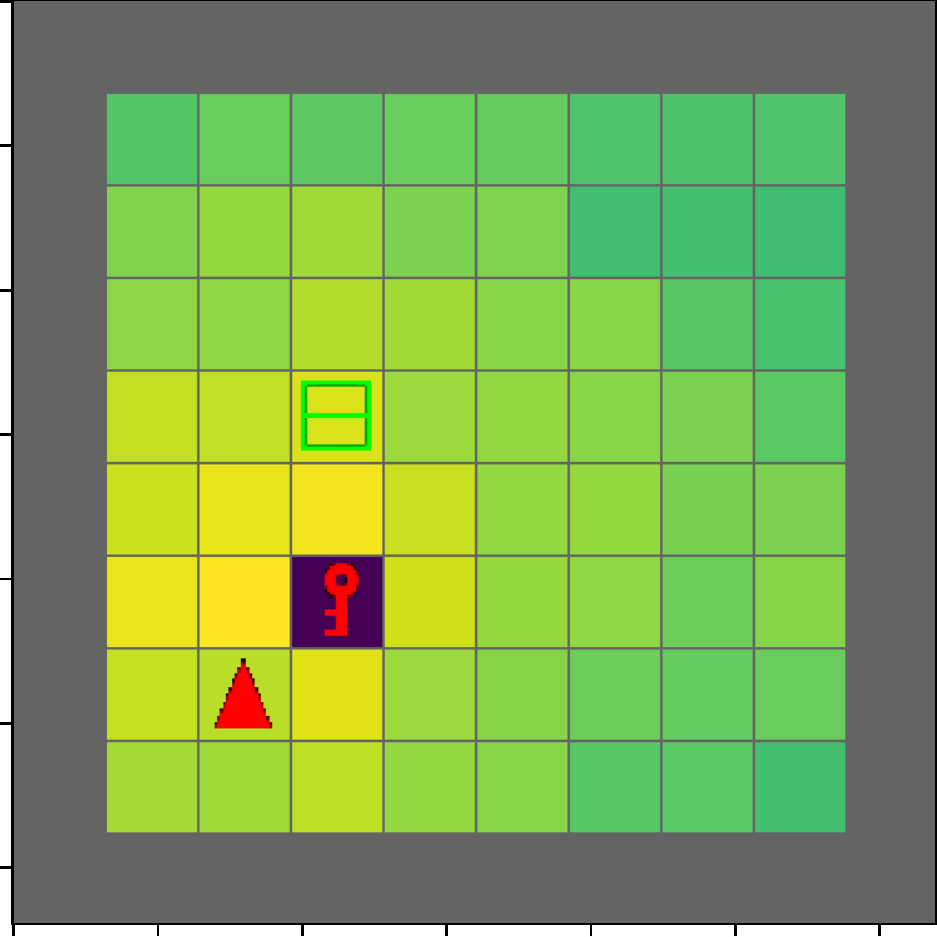}}
\raisebox{-.5\height}{\includegraphics[scale=0.23]{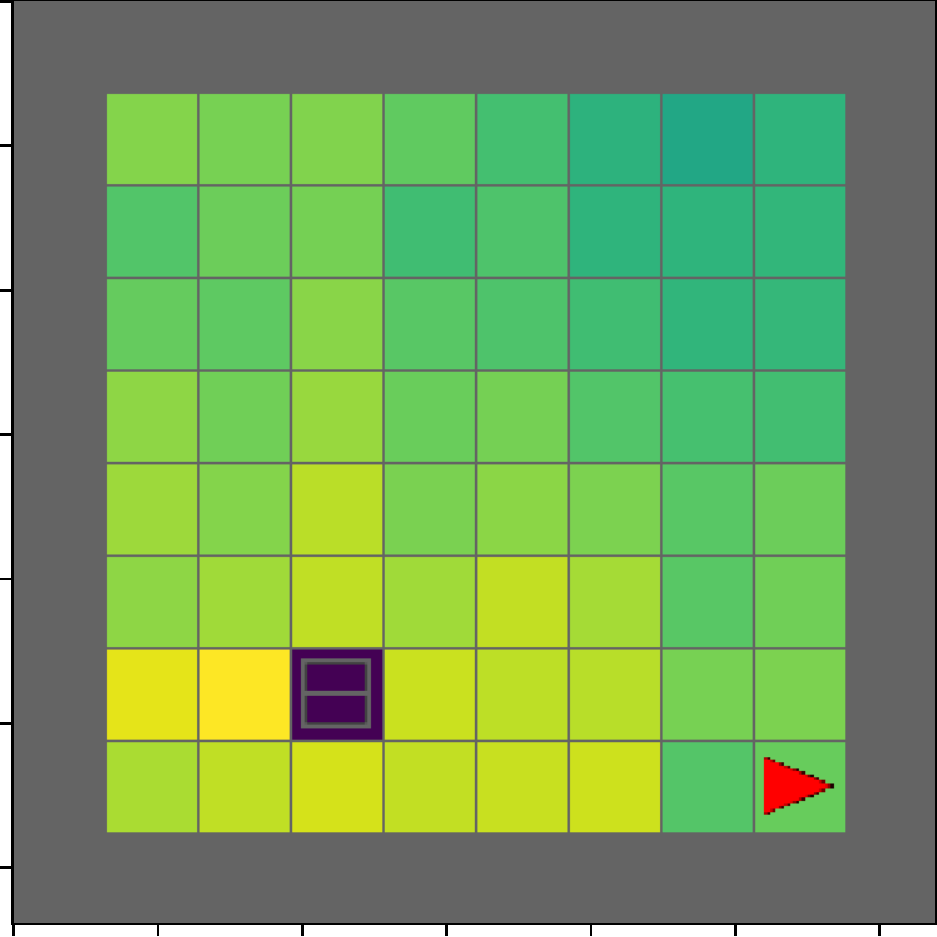}}
\raisebox{-.5\height}{\includegraphics[scale=0.23]{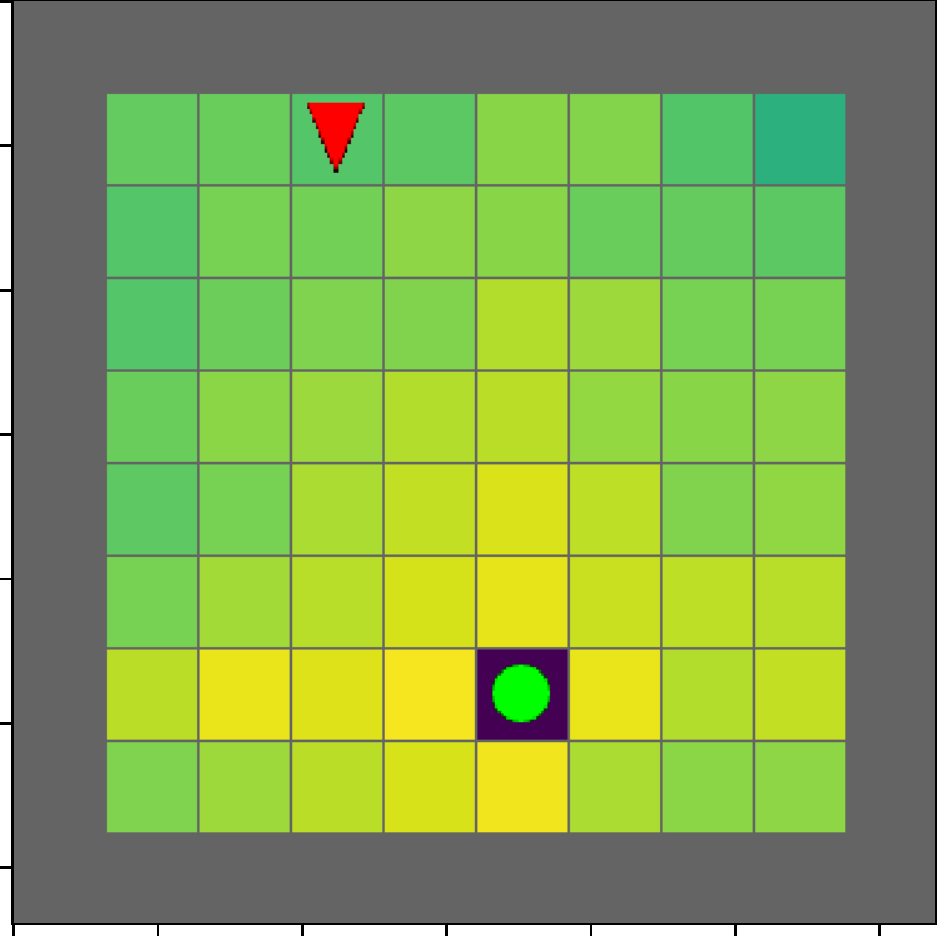}}
\raisebox{-.5\height}{\includegraphics[width=0.03\textwidth]{figures/just_colorbar_vertical.png}}\\
\\
\vspace{-0.2cm} \centerline{Affordances} \\
\hspace{-0.2cm}\subfloat[pick-purple\label{fig:test}]{\raisebox{-.5\height}{\includegraphics[scale=0.23]{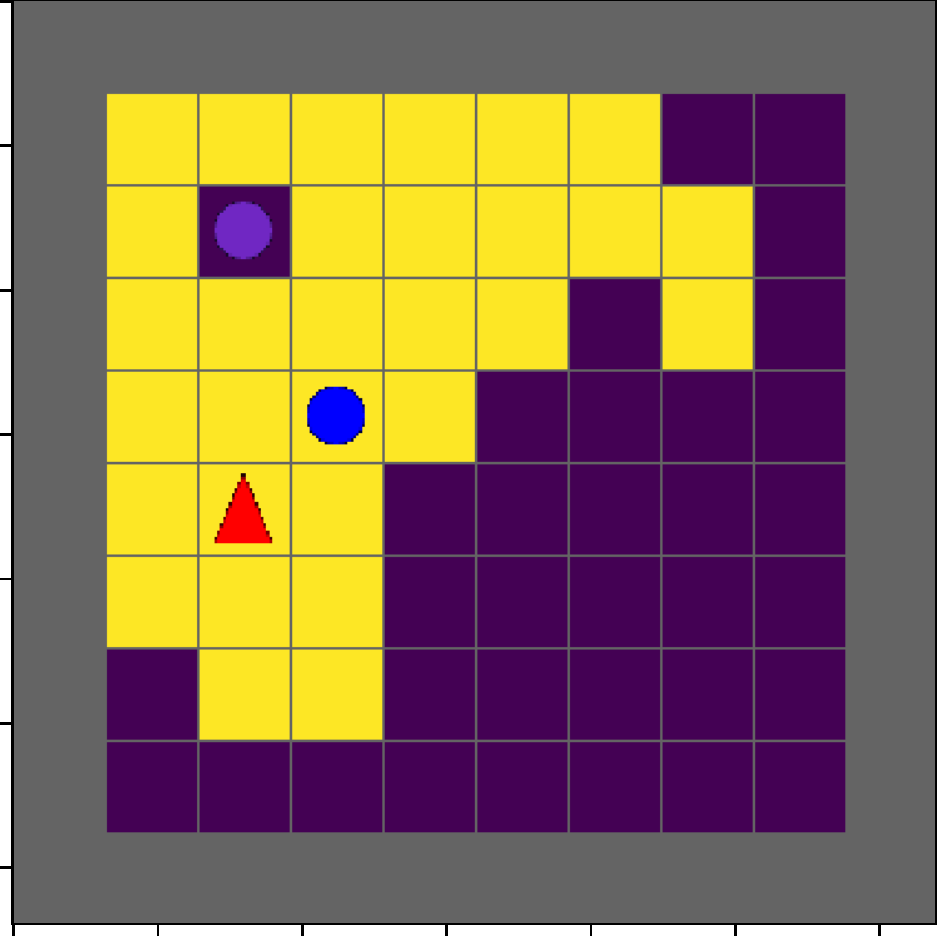}}} \hspace{0.00mm}
\subfloat[pick-blue-obj\label{fig:test}]{\raisebox{-.5\height}{\includegraphics[scale=0.23]{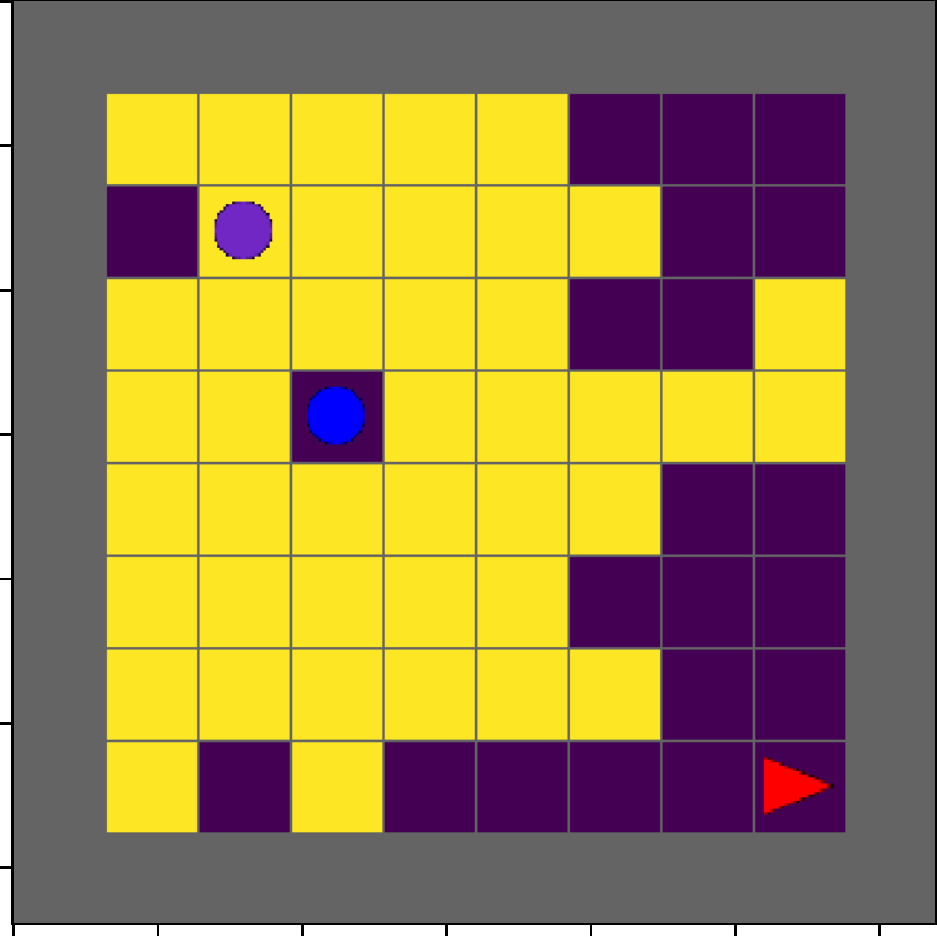}}} \hspace{0.00mm}
\subfloat[open-door\label{fig:test}]{\raisebox{-.5\height}{\includegraphics[scale=0.23]{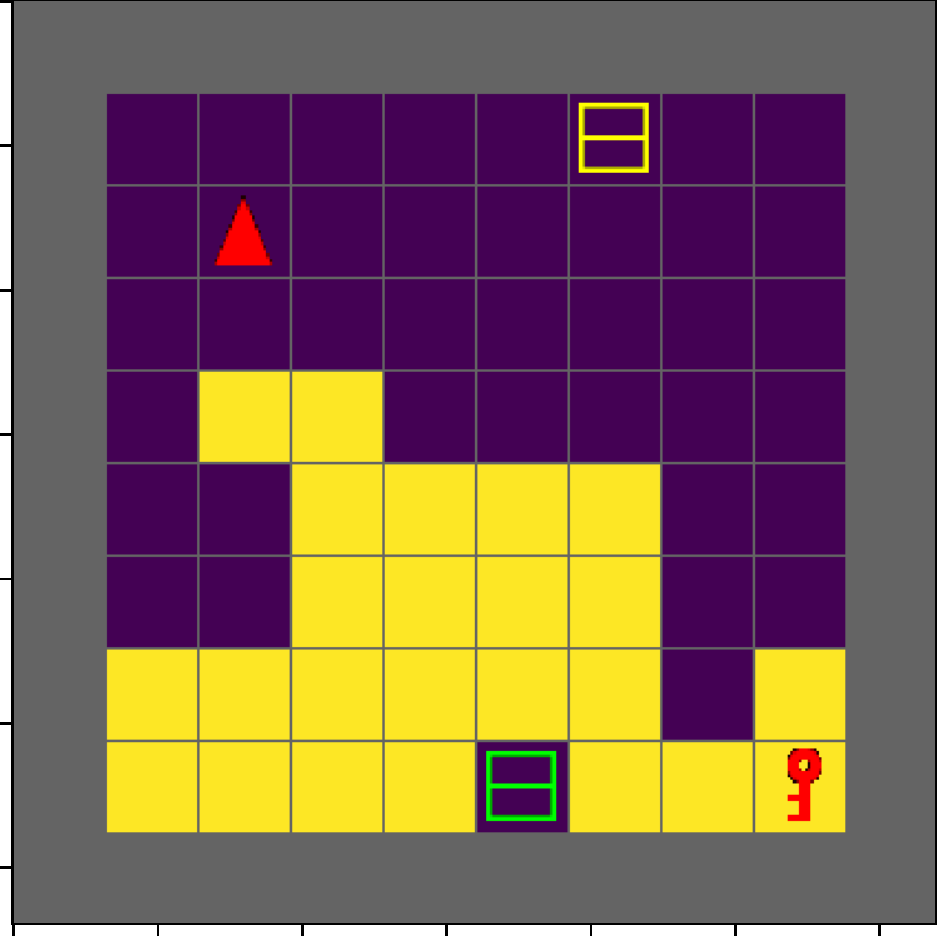}}} \hspace{0.00mm}
\subfloat[pickup-key\label{fig:test}]{\raisebox{-.5\height}{\includegraphics[scale=0.23]{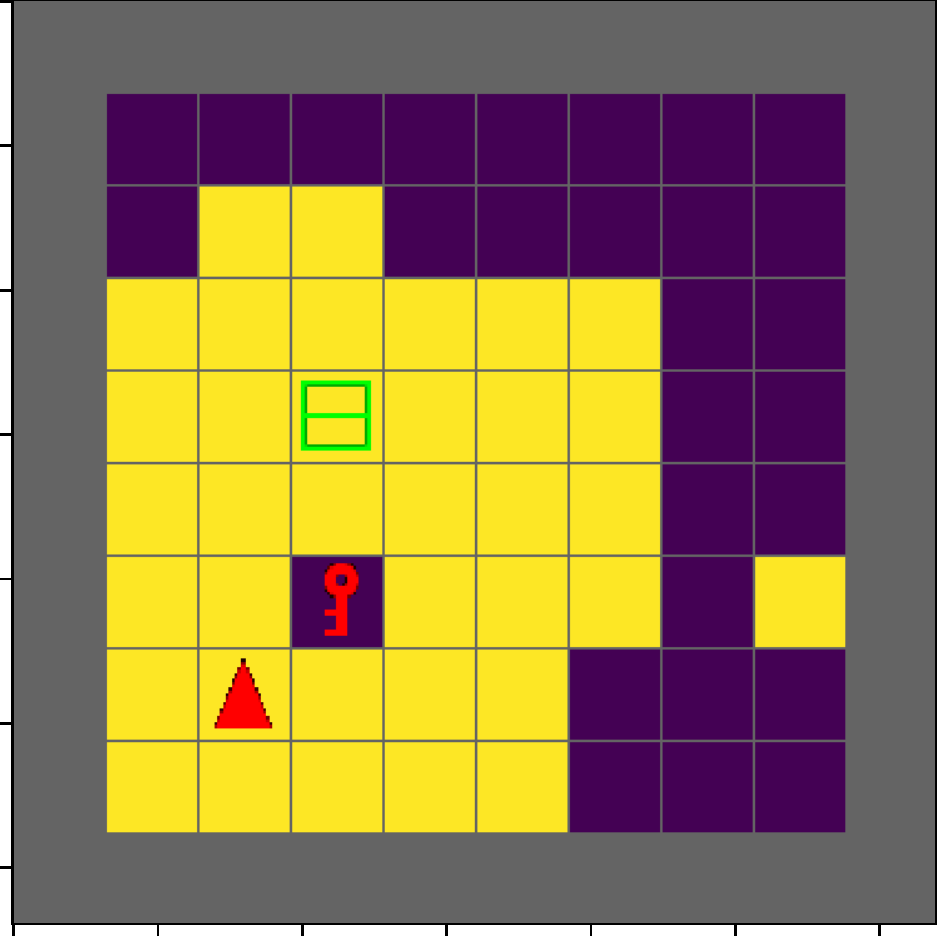}}} \hspace{0.00mm}
\subfloat[open-door\label{fig:test}]{\raisebox{-.5\height}{\includegraphics[scale=0.23]{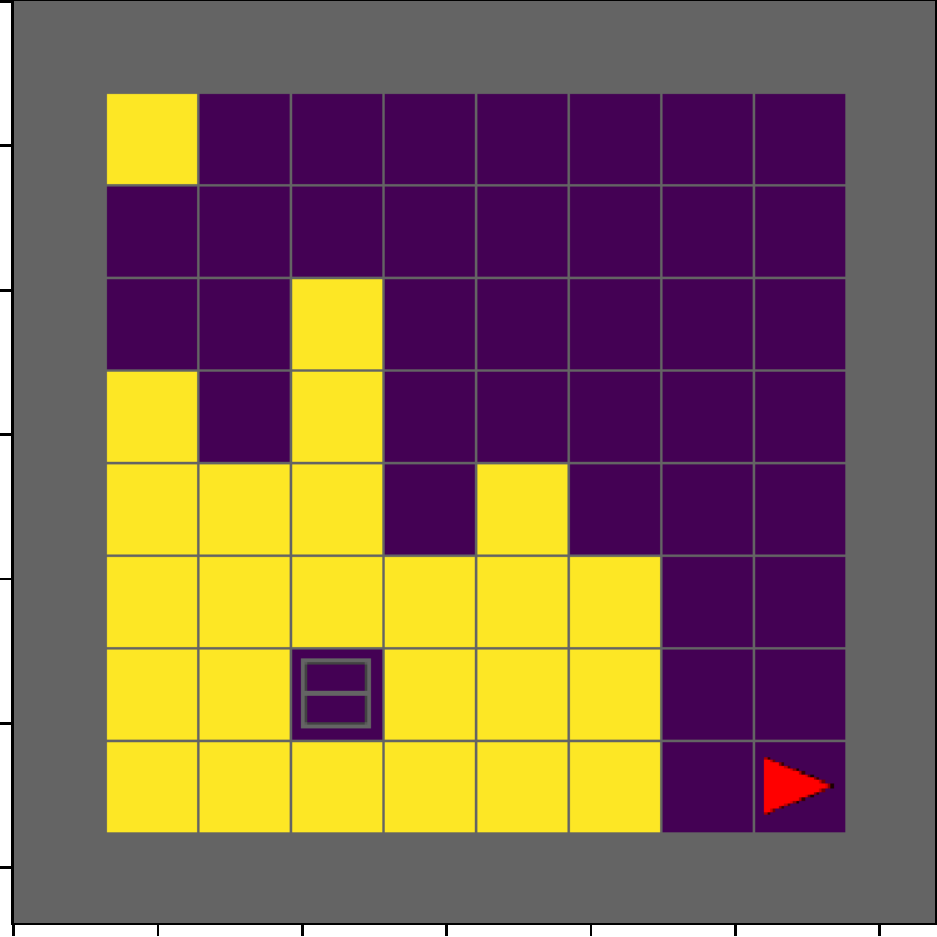}}} \hspace{0.00mm}
\subfloat[pick-green-obj\label{fig:test}]{\raisebox{-.5\height}{\includegraphics[scale=0.23]{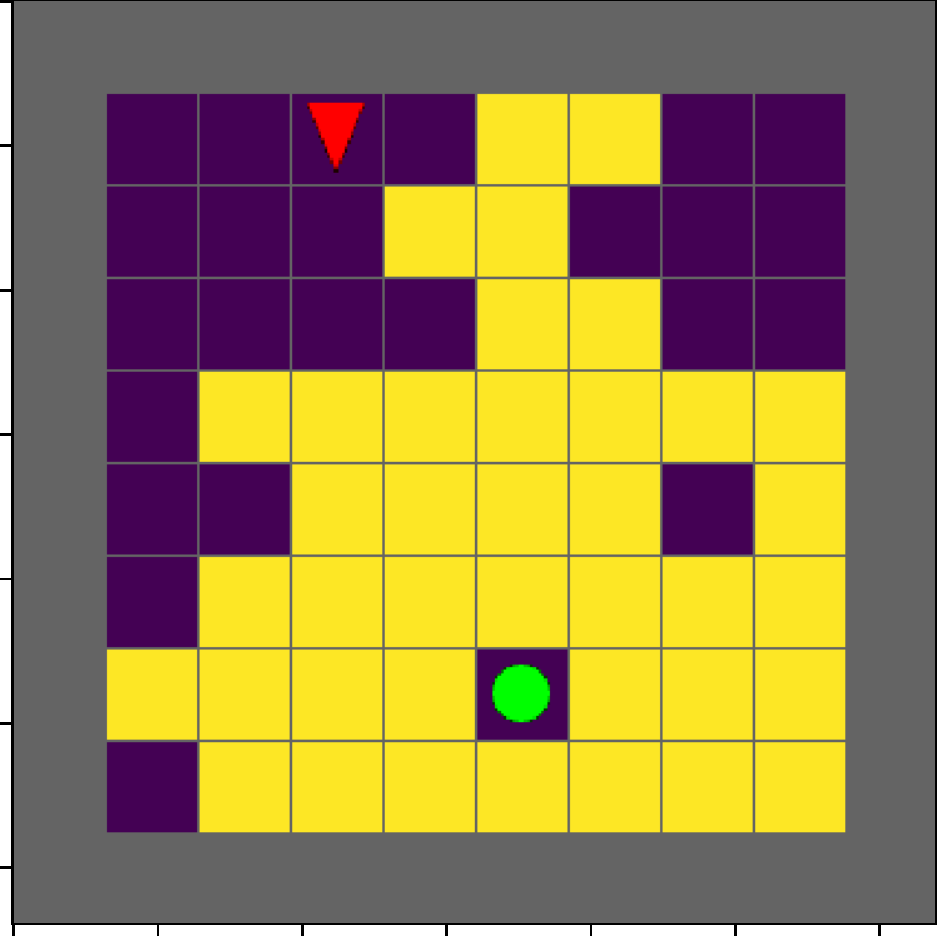}}} \hspace{0.00mm}

\end{tabular}
    \caption{Qualitative Analysis shows predictions from $\texttt{IC}^{net}$ for different subgoal intents in top row. Agent is denoted by the red triangle. Bottom row depicts option affordances obtained by a hard thresholding on $\texttt{IC}^{net}$ predictions.
    \label{figapp:qualitativeanalysis}}
    \vskip -0.2in

\end{figure}

We now show additional results in Fig.~\ref{figapp:qualitativeanalysis}, depicting the discounted intent completion network predictions and corresponding affordances derived by means of a \textit{hard}-attention. Higher values are denoted in brighter yellow while the lower ones in darker shades of green and blue. States are sampled from the environment with 4 tasks and 3 objects per task.

\subsection{Additional Experiments}
\label{sec:appendix-additionalexperiments}

\subsubsection{Analysis of $\texttt{IC}^{net}$ precision while training} 
\label{sec:appendix-icprecisionwhiletraining}
\begin{figure*}[h]
    \centering
    \includegraphics[width=0.5\textwidth]{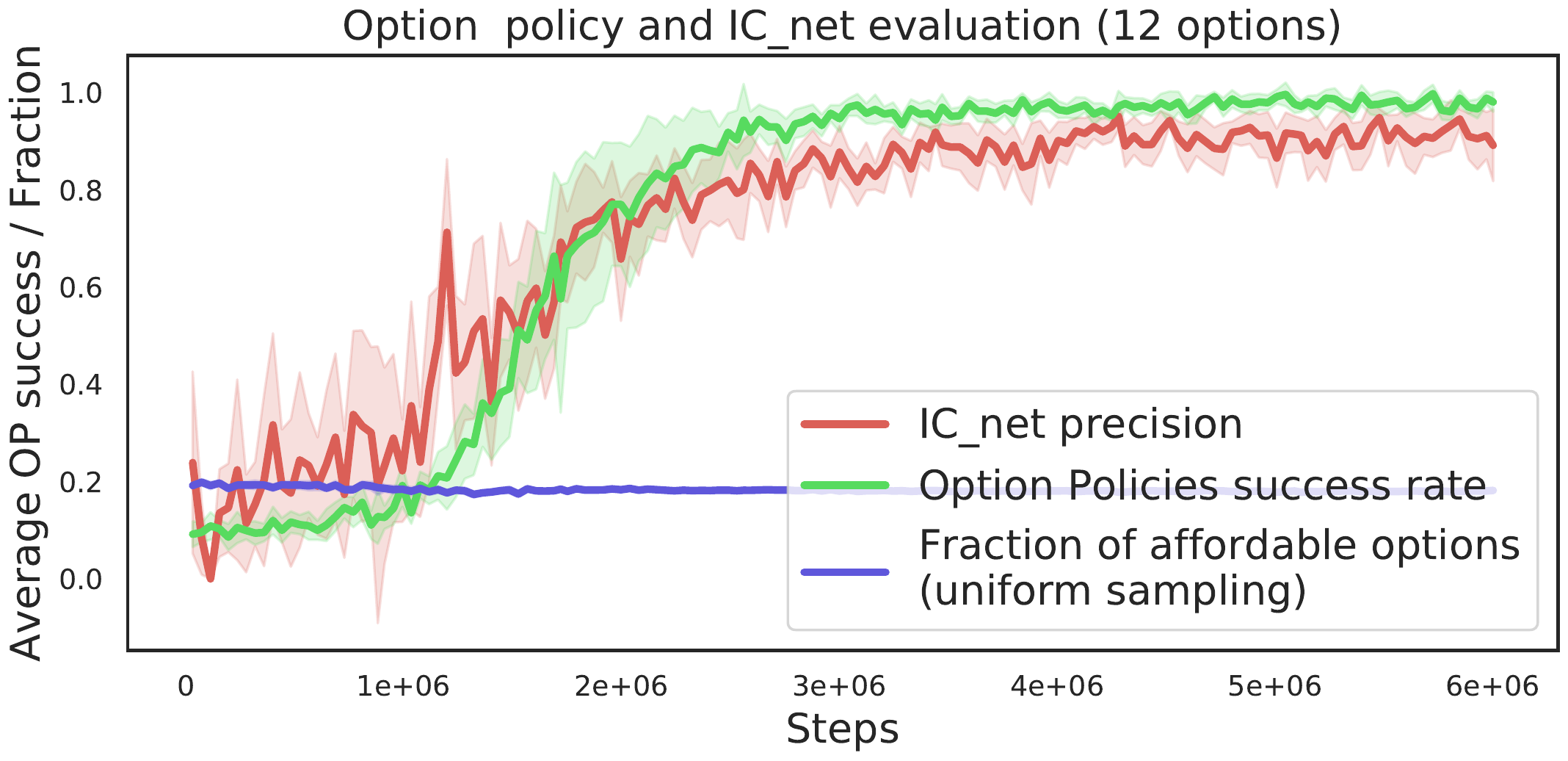}    \caption{\label{fig:ICnetversusOptionPoliciesLearning} Option success compared to the quality (precision) of the $\texttt{IC}^{net}$ from which options are sampled to collect trajectories during pre-training (Environment with 4 tasks and 3 objects per task). The blue line depicts the fraction of affordable options if sampling randomly from all environment options (12 in this experiment). The lines depict the average of 10 runs and $95\%$ confidence intervals.}
\end{figure*}

We provide in Fig.~\ref{fig:ICnetversusOptionPoliciesLearning} an additional investigation of the intent completion network's ($\texttt{IC}^{net}$) precision during training. We report the fraction of correct positive predictions made as compared to the success of option policies that use these networks for sampling affordable options during pre-training (as described in Alg.~\ref{alg:affordanceawareoptions-modelfree}). 

We find that for the same training step, the precision of the $\texttt{IC}^{net}$ is initially better. This results in skewing the option training data distribution towards option policies which are affordable - thus, quicker learning of the intent completion network allows narrowing the option set to fewer but more useful choices. This in return speeds up the option learning process as shown both here and in Fig.~\ref{fig:sampling234x3smdpvaluelearning} of the paper.

\subsubsection{Pre-training with Soft-Attention} 
\begin{figure*}[h]
    \centering
    {\includegraphics[width=0.82\textwidth]{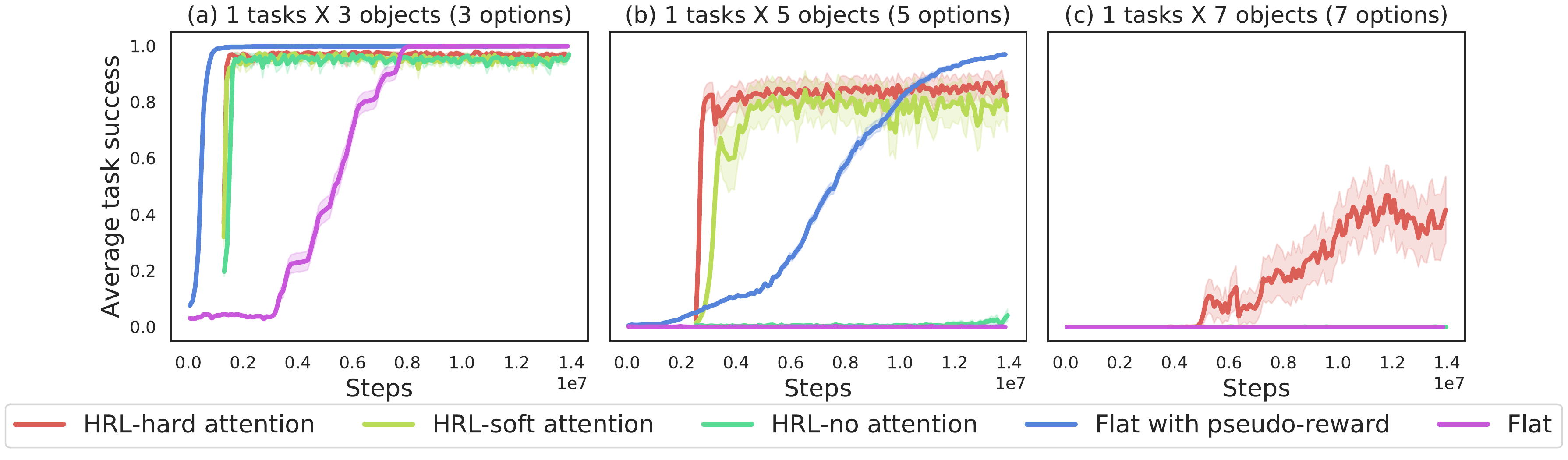}}
    \vskip -0.1in
    \caption{\label{figapp:1x357smdpvaluelearning} Long Horizon Sparse Reward Setting. Leveraging pre-trained options from \textbf{soft-attention during data generation} in downstream tasks with increasing length of the horizon from left to right. \textit{hard}-attention is able to cope better in learning speed and performance in more challenging scenarios. The lines depict the average of 10 runs and $95\%$ confidence intervals.}
\end{figure*}

\begin{figure*}[h]
    \centering
    \includegraphics[width=0.70\textwidth]{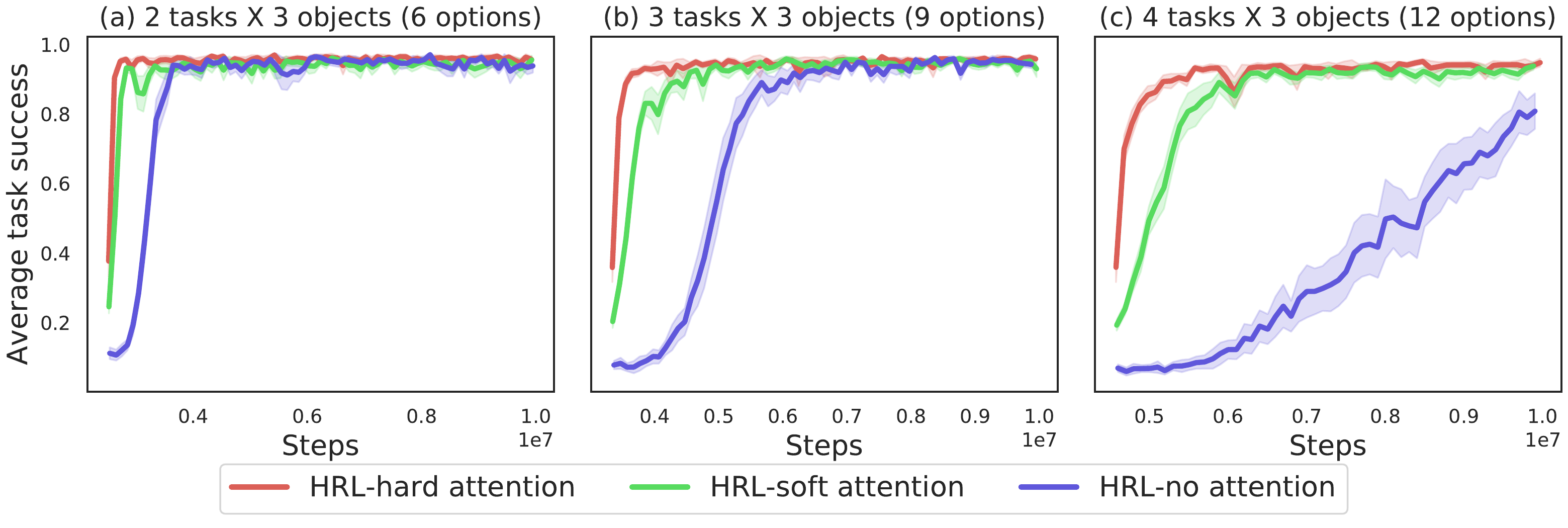}
    \vskip -0.1in
    \caption{\label{figapp:234x3smdpvaluelearning} Increasing number of choices. Leveraging pre-trained options from \textbf{soft-attention during data generation} in the downstream tasks. With growing set of choices, \textit{hard}-attention offers a more sample efficient approach by restricting to much fewer but useful choices. The lines depict the average of 10 runs and $95\%$ confidence intervals.}
\end{figure*}

We perform additional experiments to show the performance on downstream tasks when leveraging options which have been pre-trained with soft-attention as sampling technique for data generation. We repeated the experiment setup in Sec.~\ref{sec:smdpaluelearning}. Interestingly, we notice that even when using soft-attention during exploration, using hard-attention results in fewer but useful choices in both long-horizon sparse-reward tasks (see Fig.~\ref{figapp:1x357smdpvaluelearning}) and with increasing number of choices (see Fig.~\ref{figapp:234x3smdpvaluelearning}).

\subsubsection{Transition from Soft-to-Hard Attention.}
\label{sec:appendix-softtohard}
\begin{figure}[h!]
    \centering
    {\includegraphics[width=0.43\textwidth]{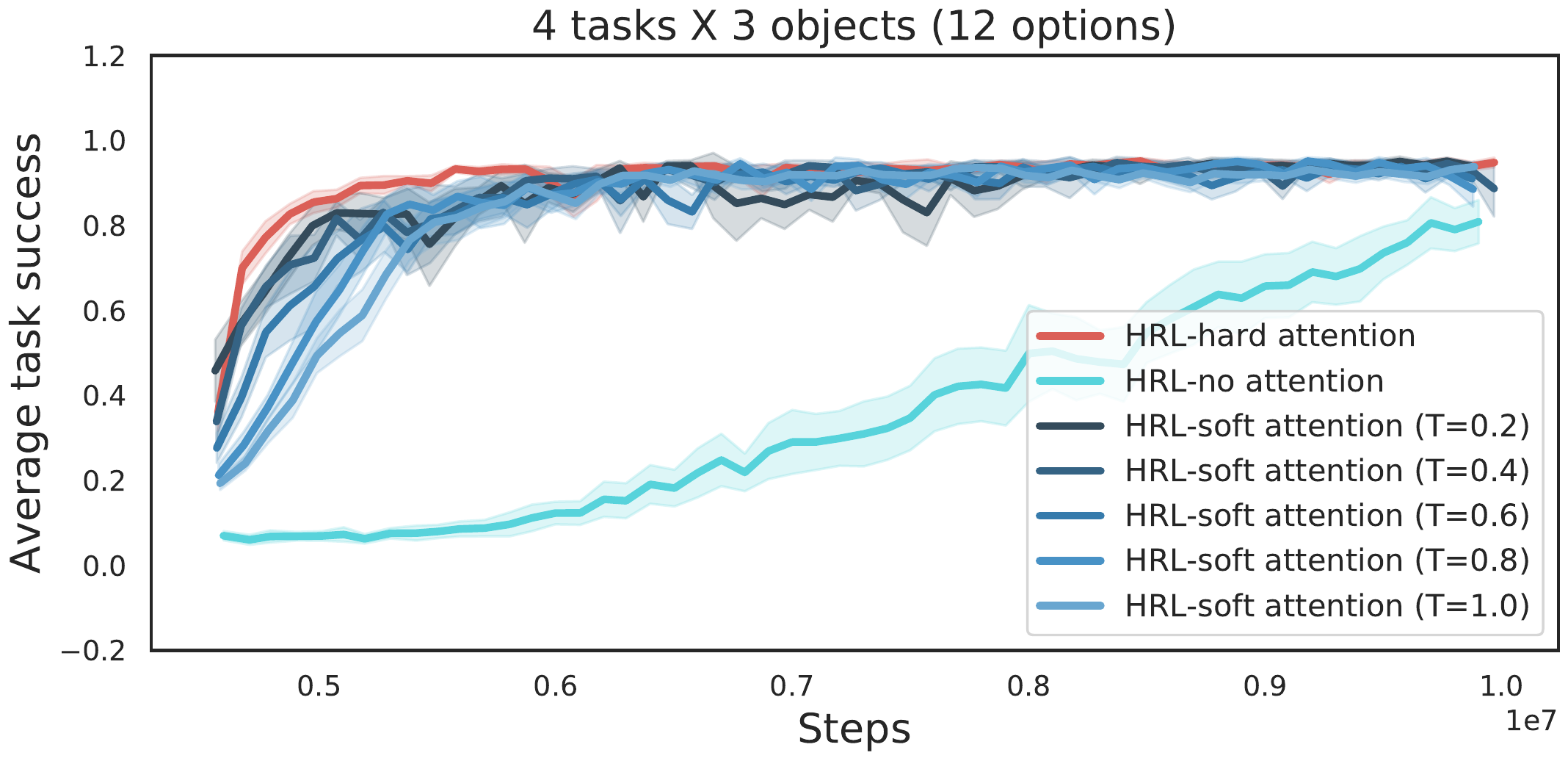}
    }
    \vskip -0.1in
    \caption{\label{figapp:interesttempvariation}\textbf{Soft-attention temperature analysis} depicts how soft-attention can be adapted using the temperature parameter of the softmax in order to close the gap of training advantage between the agents HRL-soft attention (T=1.0) and HRL-hard attention. The plot represents downstream task success rate while training different HRL agents, with pre-trained options, and with different attention mechanisms (including different values of the temperature used for soft-attention). The lines depict the average of 10 runs and $95\%$ confidence intervals.}
\end{figure}

In this work, we used the default softmax function to obtain the soft attention (see Alg~\ref{alg:affordanceawareoptions-modelfree}). However, if we consider softmax with temperature function, it facilitates transitioning from soft-attention to hard attention as shown here in Fig~\ref{figapp:interesttempvariation}. We generate here the same setting as Fig.~\ref{figapp:234x3smdpvaluelearning}(c) (right). Specifically, we plot here the default baselines that we have seen so far i.e. HRL-soft attention with temperature $1.0$, HRL-no attention, and HRL-hard attention. Next, we use the softmax with different temperature values $0.2, 0.4, 0.6, 0.8$ and show in Fig~\ref{figapp:interesttempvariation} that the temperature parameter allows us to transition from soft-to-hard attention.

\subsubsection{FetchReach: Additional Experiments}
\label{sec:appendix-fetchadditionalexperiments}

\begin{figure*}[htbp]
    \centering
    \subfloat[Pre-training \label{fig:fetch2x2pretraining}]{\includegraphics[width=0.35\textwidth]{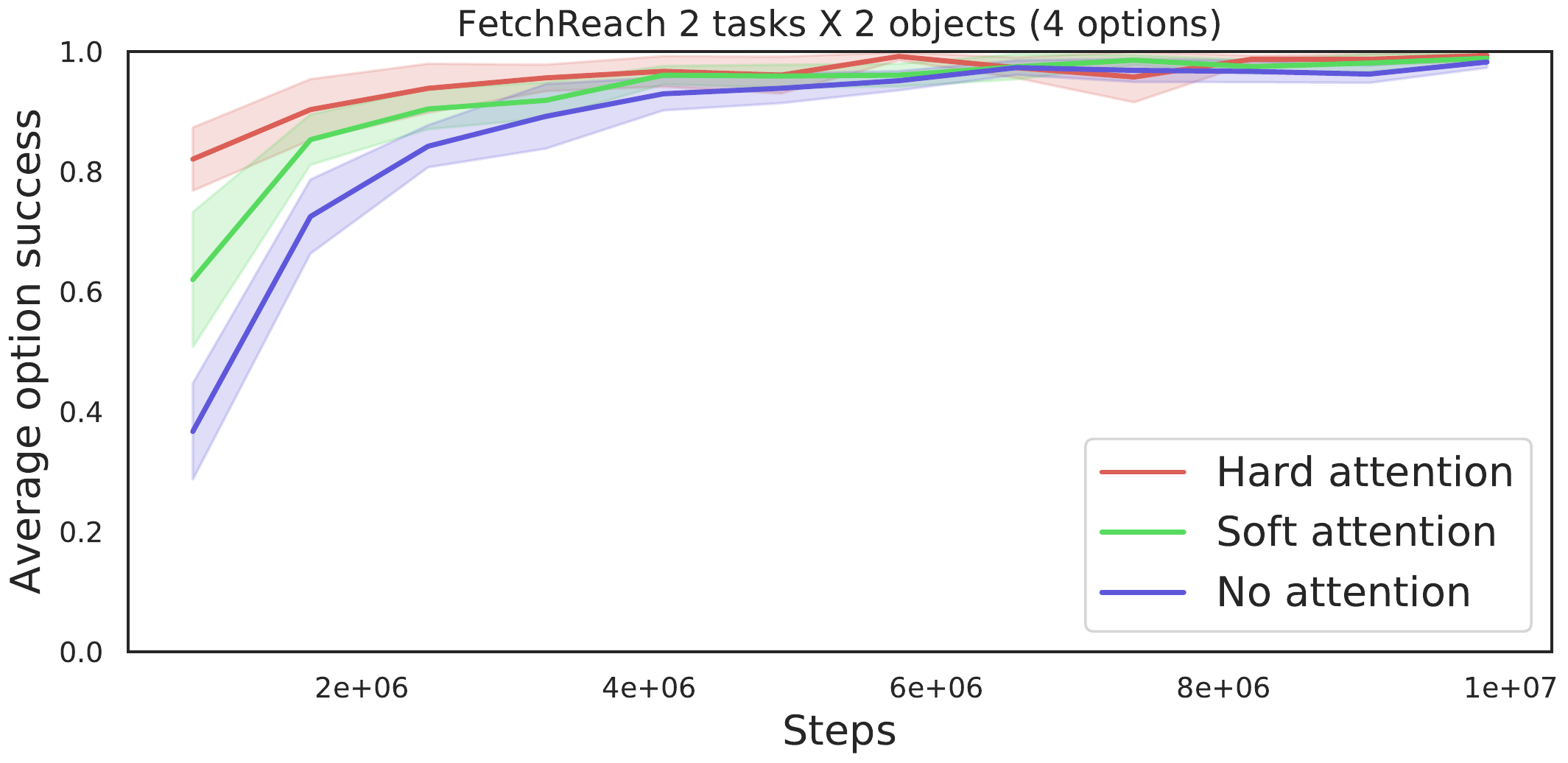}} 
    \subfloat[Downstream  Tasks\label{fig:fetch2x2downstream}]{\includegraphics[width=0.35\textwidth]{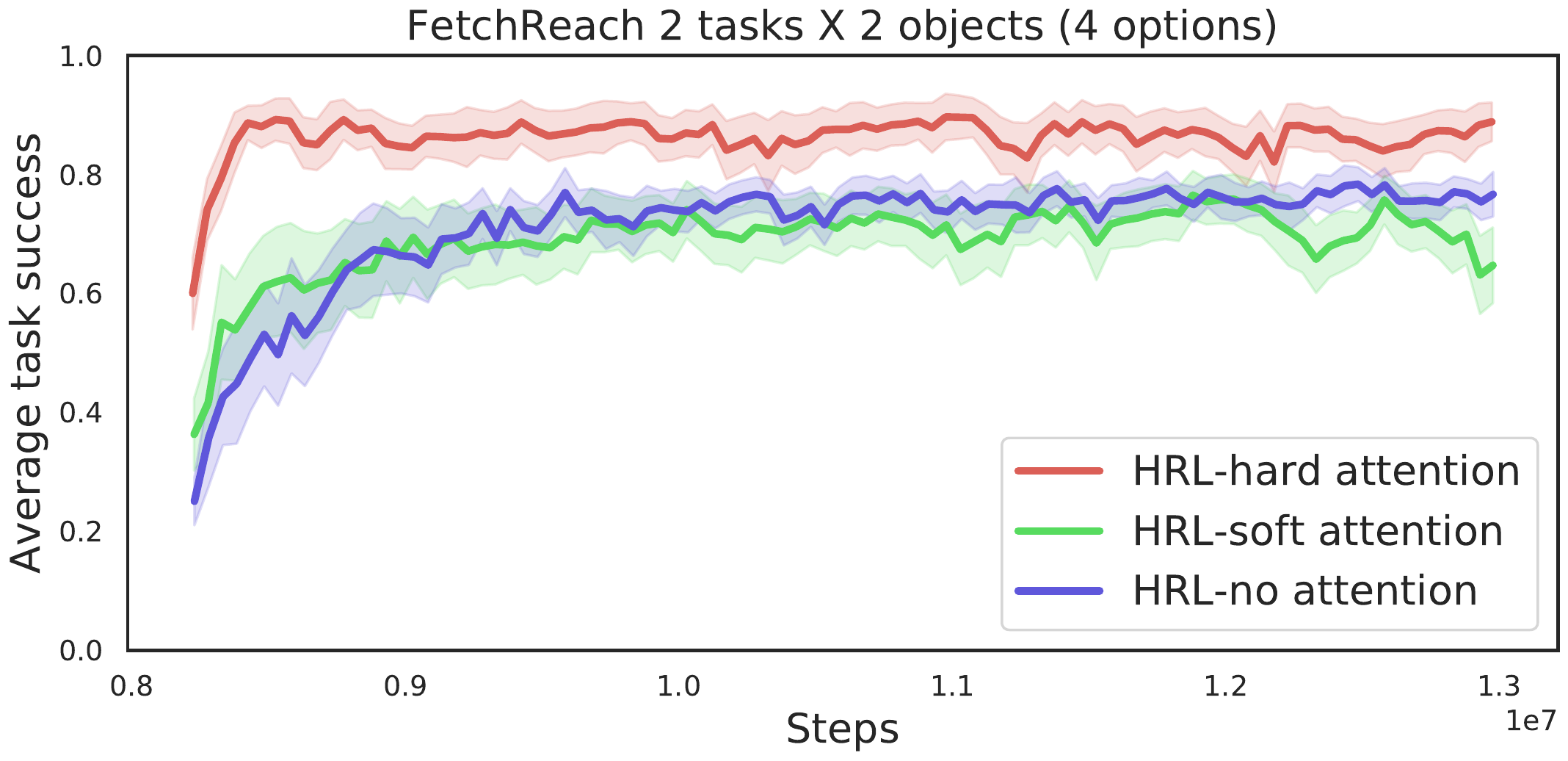}} \hspace{2mm}
    \subfloat[Pre-training \label{fig:fetch3x2pretraining}]{\includegraphics[width=0.35\textwidth]{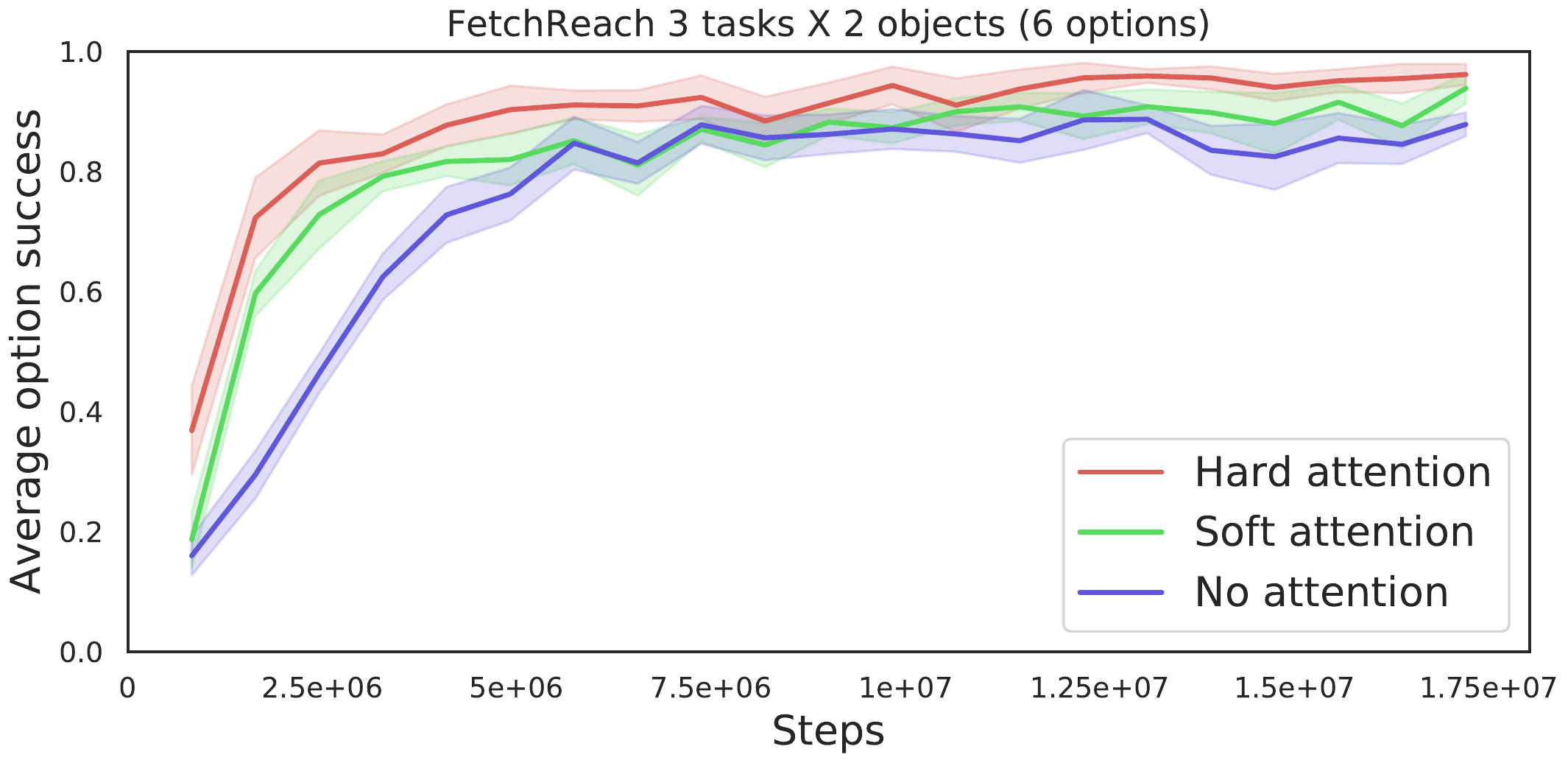}} 
    \subfloat[Downstream  Tasks\label{fig:fetch3x2downstream}]{\includegraphics[width=0.35\textwidth]{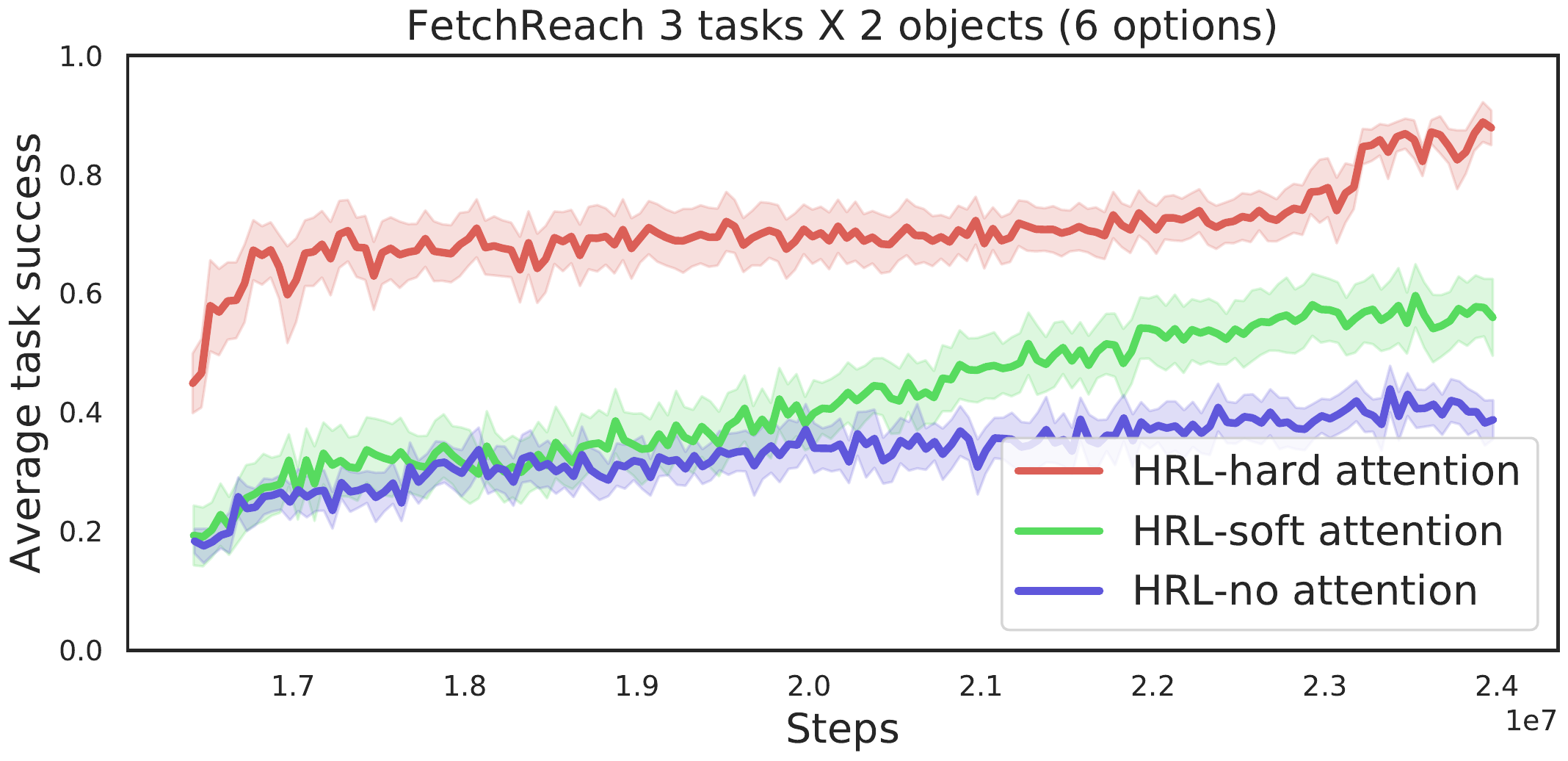}}
    \caption{\label{fig:fetch46options} Experiments for 4 and 6 options in continuous space environment. The lines depict the average of 10 runs and $95\%$ confidence intervals.}
\end{figure*}

Next, we present additional plots for Fetch Reach experiments for $4$ and $6$ options. Figures.~\ref{fig:fetch2x2pretraining} and ~\ref{fig:fetch3x2pretraining} depict option policy evaluation during pre-training with various attention types. It is observed that hard-attention has a bigger impact when more choices are to be considered i.e. $6$ options, especially compared with the no attention setup.

Besides, Figure.~\ref{fig:fetch2x2downstream} and ~\ref{fig:fetch3x2downstream} depict the average task success during planning with learned options in the downstream tasks. Learned options are loaded from checkpoints for 4/6 options at $8.19$M and $16.38$M frames. We note that during the case of $4$ options, despite the similar option quality (Fig.~\ref{fig:fetch2x2pretraining}), the HRL agent with hard agent outperforms both soft HRL agent as well as no-attention HRL agent. More significantly, this gap increases when more choices are to be considered i.e $6$ options. This result further demonstrates the paradox of choices i.e. fewer but more meaningful choices are better.

\subsection{Reproducibility}
We follow the reproducibility checklist by \cite{joelle2019} to ensure this research is reproducible. For all algorithms presented, we include a clear description of the algorithm and source code is included with these supplementary materials. For all figures that present empirical results, we include: the empirical details of how the experiments were run, a clear definition of the specific measure or statistics used to report results, and a description of results with the confidence intervals in all cases. All figures with the returns show the confidence intervals across multiple independent random seeds. Source code for the Minigrid experiments is attached in the supplementary material. We will release complete source code including Fetch experiments upon completion of the review period.

\null
\vfill

\end{document}